\def\year{2020}\relax
\documentclass[letterpaper]{article} 
\usepackage{aaai20}  
\usepackage{times}  
\usepackage{helvet} 
\usepackage{courier}  
\usepackage[hyphens]{url}  
\usepackage{graphicx} 
\usepackage{graphicx}  
\title{Stochastic Approximate Gradient Descent via the Langevin Algorithm}
\author{Yixuan Qiu\textsuperscript{\rm 1} and Xiao Wang\textsuperscript{\rm 2}\\
\textsuperscript{\rm 1}Department of Statistics and Data Science, Carnegie Mellon University, yixuanq@andrew.cmu.edu\\
\textsuperscript{\rm 2}Department of Statistics, Purdue University University, wangxiao@purdue.edu
}

\usepackage{array}
\usepackage{booktabs}
\usepackage{mathrsfs}
\usepackage{mathtools}
\usepackage{multirow}
\usepackage{amsmath}
\usepackage{amsthm}
\usepackage{amssymb}
\usepackage{subcaption}
\usepackage[ruled,vlined]{algorithm2e}

\providecommand{\tabularnewline}{\\}
\providecommand{\assumptionname}{Assumption}
\providecommand{\corollaryname}{Corollary}
\providecommand{\theoremname}{Theorem}
\providecommand{\lemmaname}{Lemma}

\theoremstyle{plain}
\newtheorem{assumption}{\protect\assumptionname}
\theoremstyle{plain}
\newtheorem{thm}{\protect\theoremname}
\theoremstyle{plain}
\newtheorem{cor}{\protect\corollaryname}
\newtheorem{lem}{\protect\lemmaname}
\theoremstyle{plain}

\newcommand{\citet}[1]{\citeauthor{#1}~\shortcite{#1}}
\newcommand{\citep}{\cite}
\newcommand{\citealp}[1]{\citeauthor{#1}~\citeyear{#1}}

\makeatletter
\newcommand{\@BIBLABEL}{\@emptybiblabel}
\newcommand{\@emptybiblabel}[1]{}
\makeatother
\usepackage[unicode=true,
 bookmarks=true,bookmarksnumbered=false,bookmarksopen=false,
 breaklinks=true,pdfborder={0 0 0},pdfborderstyle={},backref=false,colorlinks=true,
 citecolor=blue]
 {hyperref}

\begin{document}
\maketitle
\begin{abstract}
We introduce a novel and efficient algorithm called the stochastic
approximate gradient descent (SAGD), as an alternative to the stochastic
gradient descent for cases where unbiased stochastic gradients cannot
be trivially obtained. Traditional methods for such problems rely
on general-purpose sampling techniques such as Markov chain Monte
Carlo, which typically requires manual intervention for tuning parameters
and does not work efficiently in practice. Instead, SAGD makes use
of the Langevin algorithm to construct stochastic gradients that are
biased in finite steps but accurate asymptotically, enabling us to
theoretically establish the convergence guarantee for SAGD. Inspired
by our theoretical analysis, we also provide useful guidelines for
its practical implementation. Finally, we show that SAGD performs
well experimentally in popular statistical and machine learning problems
such as the expectation-maximization algorithm and the variational
autoencoders.
\end{abstract}

\section{Introduction}

The stochastic gradient descent method (SGD, \citealp{bottou2010large}; \citealp{bottou2018optimization})
is one of the most popular and widely-used optimization techniques
in large-scale machine learning problems. In many cases, the objective
function one needs to optimize can be written as an expectation, $F(\theta)=\mathbb{E}[f(\theta;\xi)]$,
over some random variable $\xi\in\mathbb{R}^{r}$ whose distribution
is independent of the parameter vector $\theta\in\Theta\subset\mathbb{R}^{p}$.
Under very mild regularity conditions, the true gradient of $F(\theta)$
is also an expectation, obtained as $g(\theta)\coloneqq\nabla F(\theta)=\mathbb{E}[\nabla f(\theta;\xi)]$.
When the computational cost of $g(\theta)$ is massive, SGD makes
use of the stochastic gradient, denoted by $\tilde{g}(\theta)$, to
update the parameter vector. It has been well studied that by appropriately
choosing the step sizes, SGD has good convergence properties \cite{robbins1951stochastic}.
As an important special case, SGD is frequently used in the scenario
where $F(\theta)$ is an average over the data points, $F(\theta)=n^{-1}\sum_{i=1}^{n}f(\theta;X_{i})$.
If data $X_{1},\ldots,X_{n}$ are assumed to be independent and identically
distributed, then an unbiased stochastic gradient can be trivially
obtained as $\tilde{g}(\theta)=\nabla f(\theta;X_{I})$, where $I$
follows a uniform distribution on $\{1,2,\ldots,n\}$.

However, there are a much broader class of problems where $\xi$ follows
a general probability distribution $\pi(\xi)$. Unlike the previous
simple scenario, in many cases an unbiased stochastic gradient cannot
be easily obtained due to the complexity of $\pi(\xi)$. If $\pi(\xi)$
is beyond the scope of standard distribution families, then some general-purpose
sampling techniques such as Markov chain Monte Carlo (MCMC, \citealp{gilks1995markov};
\citealp{metropolis1953equation}; \citealp{hastings1970monte}; \citealp{geman1984gibbs}; \citealp{brooks2011handbook})
have to be adopted, which can be quite slow in practice.

In this article, we propose a novel and efficient algorithm called
the stochastic approximate gradient descent (SAGD), as an alternative
to SGD for cases where unbiased stochastic gradients cannot be trivially
computed. The key idea of SAGD is to construct the stochastic gradient
using the Langevin algorithm \cite{roberts1996exponential,roberts2002langevin,cheng2018underdamped},
a sampling method whose statistical error can be rigorously quantified.
In addition, we use an adaptive sampling scheme that allows larger
errors in the early stage of the optimization, and gradually improves
the precision as the procedure goes on.

These heuristics are formalized in the SAGD algorithm, and various
theoretical results are developed to guarantee its convergence. Moreover,
our analysis gives clear rates of the relevant hyperparameters, which
provide useful guidelines for practical implementations of SAGD. The
highlights and main contributions of this article are as follows:
\begin{itemize}
\item We develop a new computational framework for SGD problems in which
a stochastic gradient cannot be trivially obtained. The proposed SAGD
algorithm is fully automated with a solid convergence guarantee.
\item New theoretical contributions are made to the underdamped Langevin
algorithm for sampling from sophisticated distributions, which are
of interest by their own.
\item We discuss the application of the proposed SAGD framework in some
important statistical and machine learning problems, including the
expectation-maximization algorithm (EM algorithm), and the variational
autoencoders (VAE). We show that SAGD is able to automate the EM algorithm
for complex models and effectively remove the bias of VAE.
\end{itemize}
\textbf{Notation:} Throughout this article we adopt the following
notation. Let $\mathbb{R}^{r}$ be the $r$-dimensional Euclidean
space with the inner product $\langle\cdot,\cdot\rangle$ and norm
$\Vert\cdot\Vert$. For matrices and higher-order tensors, $\Vert\cdot\Vert$
denotes the operator norm. Let $C\subset\mathbb{R}^{r}$ be a closed
convex set, and then the notation $\mathcal{P}_{C}(x)$ means the
projection of $x\in\mathbb{R}^{r}$ onto $C$. A mapping $\phi:\mathbb{R}^{r}\to\mathbb{R}^{s}$
is said to have polynomial growth if there exist a constant $C>0$
and an integer $m\ge0$ such that $\Vert\phi(x)\Vert\le C(1+\Vert x\Vert^{m})$
for all $x\in\mathbb{R}^{r}$. The notation $\nabla^{i}\phi$ is used
to denote the $i$-th derivative of a multivariate function $\phi:\mathbb{R}^{r}\to\mathbb{R}$,
and in particular $\nabla^{0}\phi\equiv\phi$. We use $\mathscr{C}_{poly}^{m}$
to denote the space of mapping $\phi$ such that $\phi$ is $m$-times
differentiable, and $\phi$ and its derivatives have polynomial growth.
A function $\phi:\mathbb{R}^{r}\to\mathbb{R}$ is said to be $L$-Lipschitz
continuous if $|\phi(x)-\phi(y)|\le L\Vert x-y\Vert$ for all $x,y\in\mathbb{R}^{r}$.

\section{Related Work}

The two main ingredients of the proposed SAGD framework are SGD and
the Langevin algorithm. SGD has been extensively studied in the literature,
and recent research mainly focused on its acceleration, for example
variance reduction methods \cite{johnson2013accelerating,reddi2016stochastic},
adaptive step sizes \cite{duchi2011adaptive,zeiler2012adadelta},
momentum methods \cite{kingma2014adam,luo2019adabound}, etc. In this
article, the proposed SAGD framework is based on the original version
of SGD, but it can be easily adapted to those acceleration methods.

The Langevin algorithm has two variants, the \emph{overdamped} and the \emph{underdamped} versions. Most of the analysis
in literature was based on the overdamped version \cite{dalalyan2017theoretical,durmus2016high,durmus2017nonasymptotic,cheng2017convergence},
whereas some recent research suggests that the underdamped version
has faster convergence for some special classes of distributions \cite{cheng2018sharp,ma2019there}.
Due to this reason, we use the underdamped Langevin algorithm to develop
the SAGD framework. As a byproduct, we have derived new results for
the underdamped Langevin algorithm that complement prior art.

The Langevin algorithm can be compared to MCMC, as they are both useful
sampling techniques. In fact, there are MCMC algorithms derived from
the Langevin algorithm such as the Metropolis-based overdamped Langevin
algorithm \cite{roberts1996exponential} and the underdamped Langevin
MCMC \cite{cheng2018underdamped}. Nevertheless, the Langevin algorithm
has several advantages in our problem. First, the Langevin algorithm
skips the Metropolis adjustment step existing in most MCMC methods,
which saves computational time and avoids duplicated values in the
sample. Second, as our theoretical analysis shows, the Langevin algorithm
has transparent hyperparameter setting and requires less manual intervention.
The downside is the resulting bias of the Langevin algorithm, but
the theoretical analysis shows that it does not harm the convergence
of SAGD.

The idea to combine SGD with the Langevin algorithm has been seen
in articles such as \citet{xie2018cooperative} and \citet{han2019divergence},
but in these works the Langevin algorithm was merely used as an MCMC-like
sampling technique, and its statistical error and impact on the convergence
of optimization were ignored. Instead, in SAGD the two ingredients
are connected in a coherent way, with a rigorous theoretical analysis.
One recent work that is similar to SAGD is \citet{de2019efficient},
but the major difference is that they used the overdamped Langevin
algorithm for sampling, whose theoretical analysis is very different
from SAGD. Another direction of research that combines SGD and the
Langevin algorithm is the stochastic gradient Langevin dynamics (SGLD,
\citealp{welling2011bayesian}; \citealp{vollmer2016exploration}).
However, SGLD utilizes SGD to accelerate the Langevin sampling method, while our
work aims at extending SGD by using the Langevin algorithm to construct
the approximate gradient.

\section{The Underdamped Langevin Algorithm}

\label{sec:langevin}

In this section we provide some background knowledge of the underdamped
Langevin algorithm, and derive a few important results that are crucial
to the convergence of SAGD. At a high level, the underdamped Langevin
algorithm is an approach to obtaining approximate samples from a target
distribution $\pi(\xi)$. In many cases, we can only compute $\pi(\xi)$
up to some normalizing constant, \emph{i.e.}, we have access to $V(\xi)\coloneqq-\log(\pi(\xi))+C$,
where $C$ is free of $\xi$. Then the \emph{underdamped Langevin
diffusion} is defined by the following stochastic differential equation
(SDE) for $W(t)=(\xi^{\mathrm{T}}(t),\rho^{\mathrm{T}}(t))^{\mathrm{T}}\in\mathbb{R}^{2r}$
with $\xi(t),\rho(t)\in\mathbb{R}^{r}$,
\begin{align}
\mathrm{d}\xi(t) & =\rho(t)\mathrm{d}t,\label{eq:sde_xi}\\
\mathrm{d}\rho(t) & =-\gamma\rho(t)\mathrm{d}t-\nabla V(\xi(t))\mathrm{d}t+\sqrt{2\gamma}\mathrm{d}B(t),\label{eq:sde_rho}\\
\xi(0) & =\xi_{0},\ \rho(0)=\rho_{0},\ t\ge0,\nonumber 
\end{align}
where $\gamma>0$ is a fixed constant but can be chosen arbitrarily,
and $B(t)$ is an $r$-dimensional Brownian motion. Under mild conditions,
Proposition 6.1 of \citet{pavliotis2014stochastic} shows that the
invariant distribution of $W(t)$ is unique, with the density function
\[
\pi_{W}(\xi,\rho)\propto\exp\{-V(\xi)-\Vert\rho\Vert^{2}/2\}\propto\pi(\xi)\cdot\exp(-\Vert\rho\Vert^{2}/2),
\]
where $\rho$ is an auxiliary variable, and our main interest is in
$\xi$. The form of $\pi_{W}(\xi,\rho)$ indicates that $\xi$ and
$\rho$ are independent with $\xi\sim\pi(\xi)$ and $\rho\sim N(0,I_{r})$.
That is, if we can solve the SDE exactly, then $\xi$ follows the
target distribution in the long run.

However, in general the solution to \eqref{eq:sde_xi} and \eqref{eq:sde_rho}
has no closed form, so some discretization methods have to be adopted.
Consider the following discretized chain for $W_{k}=(\xi_{k}^{\mathrm{T}},\rho_{k}^{\mathrm{T}})^{\mathrm{T}}$,
$k\ge0$:
\begin{align}
\xi_{k+1} & =\xi_{k}+\delta\rho_{k},\label{eq:langevin_xi}\\
\rho_{k+1} & =(1-\gamma\delta)\rho_{k}-\delta\cdot\nabla V(\xi_{k})+\sqrt{2\gamma\delta}\eta_{k},\label{eq:langevin_rho}
\end{align}
where $\delta$ is the step size, $\{\eta_{k}\}_{k=0}^{\infty}\overset{iid}{\sim}N(0,I_{r})$,
and $\eta_{k}$ is independent of $\{W_{k}\}_{i=0}^{k-1}$. The iterations
\eqref{eq:langevin_xi} and \eqref{eq:langevin_rho} are typically
referred to as the \emph{underdamped Langevin algorithm}.

The importance and usefulness of the $\{W_{k}\}$ sample will be illustrated
in Theorem \ref{thm:control_bias_mse}. Before that we need to first
guarantee that $\{W_{k}\}$ is well defined and does not explode as
time goes on. Formally, we show that under some mild conditions, $\{W_{k}\}$
is stable in the sense that it has finite moments of any order, uniformly
in the step count $k$. The result is summarized in Theorem \ref{thm:finite_moments},
along with the assumptions we need to impose.
\begin{assumption}
\label{assu:v_boundedness}(a) $V(x)$ is bounded from below, i.e.,
$V(x)\ge\nu_{0}$ for some constant $\nu_{0}\in\mathbb{R}$ and all
$x\in\mathbb{R}^{r}$. (b) The operator norm of the second derivative
of $V$ is bounded, i.e., $\Vert\nabla^{2}V(x)\Vert\le\nu$ for some
constant $\nu>0$ and all $x\in\mathbb{R}^{r}$. (c) $V(x)\in\mathscr{C}_{poly}^{\infty}$.
\end{assumption}
For Assumption \ref{assu:v_boundedness}(a), we can assume $\nu_{0}=0$
without loss of generality. This is because we can always work on
a scale-transformation of $\xi$, $\xi'=c\xi$, resulting in a transformed
$V$, $V'(x)=V(x/c)+\log c$. In what follows we adopt this simplification,
so that we have $V(x)\ge0$.
\begin{assumption}
\label{assu:v_dissipative}There exist constants $\alpha>0$ and $0<\beta<1$
such that for all $x\in\mathbb{R}^{r}$,
\[
\frac{1}{2}\langle\nabla V(x),x\rangle\ge\beta V(x)+\gamma^{2}C_{\beta}\Vert x\Vert^{2}-\alpha,\quad C_{\beta}=\frac{\beta(2-\beta)}{8(1-\beta)}.
\]
\end{assumption}
Assumption \ref{assu:v_dissipative} is a common and standard regularity
condition on $V$ coming from \citet{mattingly2002ergodicity}. We
then have the following conclusion:
\begin{thm}
\label{thm:finite_moments}Suppose Assumptions \ref{assu:v_boundedness}
and \ref{assu:v_dissipative} hold, and choose $\delta$ small enough
such that $\delta\le\min\{1/\gamma,\gamma/(2\nu),(D+1-\sqrt{D^{2}+1})/\gamma\}$,
$D=\gamma^{4}C_{\beta}/\nu^{2}$. Then for any fixed $l>0$ and all
$k\ge0$, there exist constants $C=C(l,\delta)>0$, $\lambda=\lambda(l,\delta)>0$,
and an integer $m=m(l)>0$ such that
\[
\mathbb{E}\left(\Vert\xi_{k}\Vert^{2l}+\Vert\rho_{k}\Vert^{2l}\right)\le C\left\{ 1+\left(\Vert\xi_{0}\Vert^{m}+\Vert\rho_{0}\Vert^{m}\right)e^{-\lambda k}\right\} .
\]
\end{thm}
Next, we present the main result for the underdamped Langevin algorithm.
Let $\varphi:\mathbb{R}^{2r}\to\mathbb{R}$ be a multivariate function
with the notation $\varphi(w)\equiv\varphi(\xi,\rho)$, where $w=(\xi^{\mathrm{T}},\rho^{\mathrm{T}})^{\mathrm{T}}$.
Then define its expectation with respect to $\pi_{W}$ as $\bar{\varphi}=\mathbb{E}_{\pi_{W}}\varphi\coloneqq\int\varphi(\xi,\rho)\pi_{W}(\xi,\rho)\mathrm{d}\xi\mathrm{d}\rho$.
It is easy to see that if $\varphi(w)=\nabla f(\theta;\xi)$, then
$\bar{\varphi}=\mathbb{E}_{\pi_{W}}\varphi=\nabla F(\theta)$ is exactly
the true gradient function we are interested in. Driven by the motivation
to approximate $\bar{\varphi}$, Theorem \ref{thm:control_bias_mse}
below shows that we can construct an estimator $\hat{\varphi}$ using
the sequence $\{W_{k}\}$, where $\hat{\varphi}=K^{-1}\sum_{k=0}^{K-1}\varphi(W_{k})$.
\begin{thm}
\label{thm:control_bias_mse}Let $\varphi$, $\bar{\varphi}$ , and
$\hat{\varphi}$ be defined as above, with $\varphi\in\mathscr{C}_{poly}^{r+5}$.
Assume that the conditions in Theorem \ref{thm:finite_moments} hold.
Then there exist constants $C_{1}>0$ and $C_{2}>0$ such that for
any $\delta>0$ in the range and any integer $K>0$, we have
\begin{align*}
\left|\mathbb{E}(\hat{\varphi})-\bar{\varphi}\right| & \le C_{1}\left(\frac{1}{K\delta}+\delta\right),\\
\mathbb{E}\left[\left(\hat{\varphi}-\bar{\varphi}\right)^{2}\right] & \le C_{2}\left(\frac{1}{K\delta}+\delta^{2}\right).
\end{align*}
\end{thm}
Theorem \ref{thm:control_bias_mse} shows that in general $\hat{\varphi}$
is a biased estimator for $\bar{\varphi}$, but its bias and mean
squared error can be made arbitrarily small by appropriately choosing
the algorithm parameters $\delta$ and $K$.

Here we make a few remarks about the results in this section. Theorem
\ref{thm:finite_moments} is similar to Proposition 2.7 of \citet{kopec2015weak},
but they use the implicit Euler scheme to discretize the Langevin
SDE, which is computationally much harder. Therefore, Theorem \ref{thm:finite_moments}
is a new result for the explicit Euler scheme given by \eqref{eq:langevin_xi}
and \eqref{eq:langevin_rho}. The rates in Theorem \ref{thm:control_bias_mse}
are known results \cite{chen2015convergence}. However, in most prior
art the assumptions to make Theorem \ref{thm:control_bias_mse} hold
are highly non-trivial and very difficult to check for real machine
learning models. For example, \citet{chen2015convergence} needs to
assume that our conclusion in Theorem \ref{thm:finite_moments} holds,
along with other technical conditions. In contrast, our assumptions
are only made on the log-density function $V(\xi)$, which is the
actual model that machine learning practitioners are given. In this
sense, the results developed in this article have much broader practical
use.

The benefit of our new results is that we can easily verify the assumptions
for popular machine learning models. For example, the following corollary
justifies the use of Langevin algorithm to sample from deep generative
models (e.g. VAE). Consider a single-layer neural network $h(z)=a(Wz+b)$,
where $z\in\mathbb{R}^{r}$, $b\in\mathbb{R}^{m}$, $W\in\mathbb{R}^{m\times r}$,
and the activation function is $a(x)=\log(1+e^{x})$. Then we have
the following result.
\begin{cor}
\label{cor:nn}Assume that $Z\sim N(0,I_{r})$ and $X|\{Z=z\}\sim N(h(z),\sigma^{2}I)$,
where $\sigma^{2}$ is a constant. Let $p(z|x)$ denote the conditional
density of $Z$ given $X=x$, and then $V(z)=-\log p(z|x)$ satisfies
Assumptions \ref{assu:v_boundedness} and \ref{assu:v_dissipative}.
\end{cor}
For brevity we omit the multi-layer case, but it can be analyzed similarly.
In later part of this article we will discuss the application of SAGD
to VAE model in more details.

\section{Stochastic Approximate Gradient Descent}

With the statistical properties of the underdamped Langevin algorithm
studied in Theorem \ref{thm:control_bias_mse}, the SAGD framework
can then be readily developed. Recall that our target is to minimize
the function $F(\theta)=\mathbb{E}[f(\theta;\xi)]$, whose true gradient
$g(\theta)=\mathbb{E}[\nabla f(\theta;\xi)]$ is hard to compute exactly.
Using the technique developed in the previous section, we can construct
a stochastic gradient, $\tilde{g}(\theta)=K^{-1}\sum_{k=0}^{K-1}\nabla f(\theta;\xi_{k})$,
to approximate $g(\theta)$. Unlike most existing SGD settings, $\tilde{g}(\theta)$
is not an unbiased estimator for $g(\theta)$, as suggested by Theorem
\ref{thm:control_bias_mse}. Therefore, we refer to the optimization
method based on such a $\tilde{g}(\theta)$ as the stochastic \emph{approximate}
gradient descent. The outline of SAGD is given in Algorithm \ref{alg:sagd}.

\begin{algorithm}
\caption{\label{alg:sagd}Stochastic approximate gradient descent for minimizing
$F(\theta)=\mathbb{E}[f(\theta;\xi)]$}

\SetKwInOut{Input}{Input}
\SetKwInOut{Output}{Output}
\Input{$T$, $\{\alpha_{t}\}$, $\{\delta_{t}\}$, $\{K_{t}\}$, initial values $\theta_{0}$, $\xi_{0}$, $\rho_{0}$}
\Output{Parameter estimate for $\theta$}

\For{$t=0,1,\ldots,T-1$}{
    $\xi_{t,0}\leftarrow\xi_{0}$, $\rho_{t,0}\leftarrow\rho_{0}$\;

    \For{$k=1,2,\ldots,K_{t}-1$}{
        $\xi_{t,k+1}\leftarrow\xi_{t,k}+\delta_{t}\rho_{t,k}$\;
        Sample $\eta_{t,k}\sim N(0,I_{r})$\;
        $\rho_{t,k+1}\leftarrow(1-\gamma\delta_{t})\rho_{t,k}-\delta_{t}\cdot\nabla V(\xi_{t,k})+\sqrt{2\gamma\delta_{t}}\eta_{t,k}$\;
    }

    $\tilde{g}_{t}(\theta)\leftarrow K_{t}^{-1}\sum_{k=0}^{K_{t}-1}\nabla f(\theta;\xi_{t,k})$\;
    $\theta_{t+1}\leftarrow\mathcal{P}_{\Theta}\left(\theta_{t}-\alpha_{t}\cdot\tilde{g}_{t}(\theta_{t})\right)$\;
}

\Return{$\hat{\theta}=T^{-1}\sum_{t=1}^{T}\theta_{t}$}

\end{algorithm}

Despite the fact that $\tilde{g}(\theta)$ is a biased estimator for
the true gradient, we show that by carefully choosing the hyperparameters,
we can actually guarantee the overall convergence of SAGD. Interestingly,
the convergence rate for a convex objective function, in terms of
the number of gradient updates, is the same as the vanilla SGD method
with an order of $\mathcal{O}(1/\sqrt{T})$, as is shown in Theorem
\ref{thm:convergence_convex}.
\begin{assumption}
\label{assu:f_condition}$f(\theta;\cdot)\in\mathscr{C}_{poly}^{r+5}$
for each $\theta\in\Theta$, and there exist a constant $C>0$ and
an integer $m\ge0$ such that $\Vert\nabla^{i}f(\theta;\cdot)\Vert\le C(1+\Vert\cdot\Vert^{m})$
for all $\theta\in\Theta$ and $0\le i\le r+5$.
\end{assumption}
\begin{thm}
\label{thm:convergence_convex}Suppose that $F(\theta)$ is convex
and $L$-Lipschitz continuous in $\theta\in\Theta$, and $\Theta$
is a closed convex set with diameter $D<\infty$. Also assume that
Assumption \ref{assu:f_condition} and the conditions in Theorem \ref{thm:finite_moments}
hold. Then by choosing $\delta_{t}=C_{1}/\sqrt{t}$, $K_{t}=C_{2}t$,
and $\alpha_{t}=\alpha_{0}/\sqrt{t}$, where $C_{1},C_{2},\alpha_{0}>0$
are constants, we have $\mathbb{E}[F(\hat{\theta})]-F^{*}\le\mathcal{O}(1/\sqrt{T})$.
\end{thm}
The significance of Theorem \ref{thm:convergence_convex} is that
it provides clear rates for the hyperparameters $\delta_{t}$ and
$K_{t}$ in the sampling algorithm, which are crucial for practical
algorithm implementation but are typically missing in other MCMC-based
methods. Of course, the preservation of the SGD rate is not without
a price. Theorem \ref{thm:convergence_convex} indicates that the
number of inner iterations, \emph{i.e.}, $K_{t}$ in Algorithm \ref{alg:sagd},
needs to increase with $t$. However, the developed error bounds are
typically conservative, so for practical use, we advocate the following
techniques to speed up SAGD: (1) An educated initial value $\xi_{0}$
can be used to initialize the Langevin algorithm, for example in VAE
$\xi_{0}$ is sampled from the trained encoder; (2) A persistent Langevin
Markov chain is stored during optimization, motivated by the persistent
contrastive divergence \cite{tieleman2008training}; (3) Some advanced
gradient update schemes such as Adam can be used.

More generally, we consider objective functions that are nonconvex
but smooth, and assume that $\Theta=\mathbb{R}^{p}$. Theorem \ref{thm:convergence_nonconvex}
indicates that with a proper choice of hyperparameters, the algorithm
again has a nice convergence property.
\begin{thm}
\label{thm:convergence_nonconvex}Suppose that $g(\theta)$ is $G$-Lipschitz
continuous in $\theta$, and assume that Assumption \ref{assu:f_condition}
and the conditions in Theorem \ref{thm:finite_moments} hold. Let
$\delta_{t}=C_{1}t^{-c}$, $K_{t}=C_{2}t^{2c}$, and $\alpha_{t}=\alpha_{0}/t$
for some constants $0<\alpha_{0}<1/(2G)$ and $C_{1},C_{2},c>0$.
Then we have $\lim\inf_{t\rightarrow\infty}\mathbb{E}[\Vert g(\theta_{t})\Vert^{2}]=0$.
\end{thm}
Theorem \ref{thm:convergence_nonconvex} is an analog to Theorem 4.9
of \citet{bottou2018optimization}. It is not meant to be the strongest
conclusion, but to provide insights on the convergence property of
SAGD for nonconvex objective functions.

\section{Applications: EM Algorithm and VAE}

\label{sec:application}

\subsection{Automated EM Algorithm}

The SAGD framework is very useful for implementing an automated version
of the EM algorithm \cite{dempster1977maximum}. EM algorithm is a
powerful and indispensable tool to solve missing data problems and
latent variable models. Given the data set $X$ that follows a probability
distribution with density function $f(x;\theta)$, we are interested
in computing the maximum likelihood estimator $\hat{\theta}=\arg\max_{\theta}\,\ell(\theta;x)$
for the unknown parameter vector $\theta$, where $\ell(\theta;x)=\log[f(x;\theta)]$
is the log-likelihood function.

However, in many cases the computation of the marginal distribution
$f(x;\theta)$ is intractable, but with an additional random vector
$U$, the complete log-likelihood $L(\theta;x,u)=\log[f(x,u;\theta)]$
is simple. This phenomenon typically happens when $U$ represents missing
data or latent variables in the model. The EM algorithm computes $\hat{\theta}$
in an iterative way. Given the current value of $\theta$, denoted
by $\theta_{k}$, the EM algorithm proceeds by the following two steps:
\begin{itemize}
\item \textbf{Expectation Step (E-step)}: Compute the expected value of
$L(\theta;x,U)$ with respect to the conditional distribution of $U$
given $X=x$ under the current parameter estimate $\theta_{k}$, and
define the function $Q(\theta;\theta_{k})=\mathbb{E}_{U|X=x,\theta_{k}}[L(\theta;x,U)]$.
\item \textbf{Maximization Step (M-step)}: Update the estimate of $\theta$
by maximizing the $Q$ function: $\theta_{k+1}=\arg\max_{\theta}\,Q(\theta;\theta_{k})$.
\end{itemize}
The EM algorithm has a remarkable monotonicity property, \emph{i.e.},
the marginal log-likelihood $\ell(\theta;x)$ is always nondecreasing
on the $\{\theta_{k}\}$ sequence. Due to such nice properties, the
EM algorithm has been the standard optimization technique for Gaussian
mixture models and many other missing data models. However, one serious
problem of the EM algorithm is that the expectation defining the $Q$
function usually has no simple closed form, so the Monte Carlo EM
algorithm (MCEM, \citealp{wei1990monte}; \citealp{levine2001implementations}) proposes
to use Monte Carlo methods to approximate the expectation. Using MCMC
to approximate the expectation in the E-step is not a new idea, but
what really matters is how to properly choose the hyperparameters
to guarantee the convergence of the M-step.

In this sense, Theorem \ref{thm:convergence_convex} provides a clear
way to make the EM algorithm effectively automated. It is easy to
see that the target distribution $\pi(\xi)$ is $p(u|x;\theta_{k})$,
the conditional density of $U$ given $X=x$, which is proportional
to the joint density of $(X,U)$ under $\theta_{k}$. Therefore, we
can define $V(\xi)=-L(\theta_{k};x,\xi)$, and then apply Algorithm
\ref{alg:sagd} to directly solve the M-step, whose convergence is
readily guaranteed. Finally, one only needs to create an outer loop
to iteratively update the $\{\theta_{k}\}$ sequence, until some convergence
condition is met.

\subsection{Debiased VAE}

\label{subsec:vae_bias_correction}

The automated EM algorithm can be further used to improve the popular
VAE model \cite{kingma2014stochastic}. VAE has the same goal of seeking
the maximum likelihood estimator for $\theta$, but it uses the variational
Bayes technique to maximize a lower bound of $\ell(\theta;x)$. Let
$q(u|x)$ be any conditional density function, and then VAE maximizes
the function $\tilde{\ell}(\theta;x)$, defined by
\begin{equation}
\tilde{\ell}(\theta;x)=\mathbb{E}_{u\sim q}[\log p(x|u;\theta)]-\mathrm{KL}[q(u|x)\Vert p(u)],\label{eq:elbo}
\end{equation}
where $p(u)$ is the marginal density of $u$, and $p(x|u;\theta)$
is the conditional distribution of $X$ given $U=u$. In most VAE
settings, $U\sim N(0,I)$, and $q(u|x)$ is taken to be a normal distribution
whose mean and variance parameters are represented by a deep neural
network. VAE has been successfully applied to many problems, but its
most critical weakness is that VAE does not maximize the exact log-likelihood,
which induces a bias in the final $\theta$.

Here we show that using the SAGD framework, the bias of VAE can be
removed via an additional refining step. First, it is easy to show
that $\tilde{\ell}(\theta;x)$ has another representation, $\tilde{\ell}(\theta;x)=\ell(\theta;x)-\mathrm{KL}[q(u|x)\Vert p(u|x;\theta)]$.
That is, if the distribution $q(u|x)$ matches the true $p(u|x;\theta)$,
then $\tilde{\ell}(\theta;x)$ is the genuine log-likelihood function
$\ell(\theta;x)$. In this case, the objective function \eqref{eq:elbo}
can be optimized via an EM algorithm with a $Q$ function $Q(\theta;\theta_{k})=\mathbb{E}_{U\sim p(u|x;\theta_{k})}[\log p(x|u;\theta)+\log p(u)]$,
and we update the current parameter $\theta_{k}$ by a gradient move
\[
\theta_{k+1}=\theta_{k}+\alpha_{k}\cdot[\partial Q(\theta;\theta_{k})/\partial\theta]|_{\theta=\theta_{k}},
\]
with the true expectation replaced by the Langevin-based approximate
gradient. The consequence of this refining step is that we are now
optimizing the true log-likelihood function $\ell(\theta;x)$ instead
of the lower bound $\tilde{\ell}(\theta;x)$, and hence the bias of
VAE is removed.

We emphasize that we do not position SAGD as a replacement for VAE;
in fact, VAE is computationally more efficient and has a lower variance.
Instead, the major virtue of SAGD is its bias-correction capacity
that fixes the intrinsic gap between the evidence lower bound of VAE
and the true likelihood. Therefore, we suggest using VAE to pre-train
models, and then fine-tuning the generative network using SAGD due
to its theoretical guarantee.

\section{Numerical Experiments}

\subsection{EM Algorithm}

In this section we use numerical experiments to demonstrate the applications
of SAGD in EM algorithm and VAE as discussed in the previous section.
First consider a simple model such that the parameter estimation procedure
can be easily visualized. Assume that given latent variables $Z_{1},\ldots,Z_{n}\overset{iid}{\sim}N(0,1)$,
the data are independently generated as $X_{i}|\{Z_{1}=z_{1},\ldots,Z_{n}=z_{n}\}\sim\mathsf{Gamma}(10\cdot\sigma(a+bz_{i}))$,
where $\sigma(x)=1/(1+\exp(-x))$ is the sigmoid function, and $\mathsf{Gamma}(s)$
stands for a gamma distribution with shape parameter $s$. The target
is to estimate the unknown parameters $\theta=(a,b)$ from the observed
data $X_{1},\ldots,X_{n}$. In our simulation, the true parameters
are set to $a=2$ and $b=0.5$, and a sample size $n=100$ is used
to simulate $X_{i}$.

For a single variable pair $(x,z)$, it is easy to show that the complete
log-likelihood function is $L(\theta;x,z)=-z^{2}/2+(s-1)\log(x)-\log\{\Gamma(s)\}+C$,
where $s=10\cdot\sigma(a+bz)$, $\Gamma(\cdot)$ is the gamma function,
and $C$ is a constant. The EM algorithm is then used to solve this
problem as follows. In the $k$-th M-step, we fix parameter estimate
at $\theta_{k}=(a_{k},b_{k})$, and then optimize the objective function
$Q(\theta;\theta_{k})=\mathbb{E}_{Z|X=x,\theta_{k}}[L(\theta;x,Z)]$
using SAGD. The next $\theta$ value is set to the optimum of $Q(\theta;\theta_{k})$.

\begin{figure*}
\begin{centering}
  \begin{subfigure}[b]{0.48\textwidth}
  \centering\includegraphics[width=\textwidth]{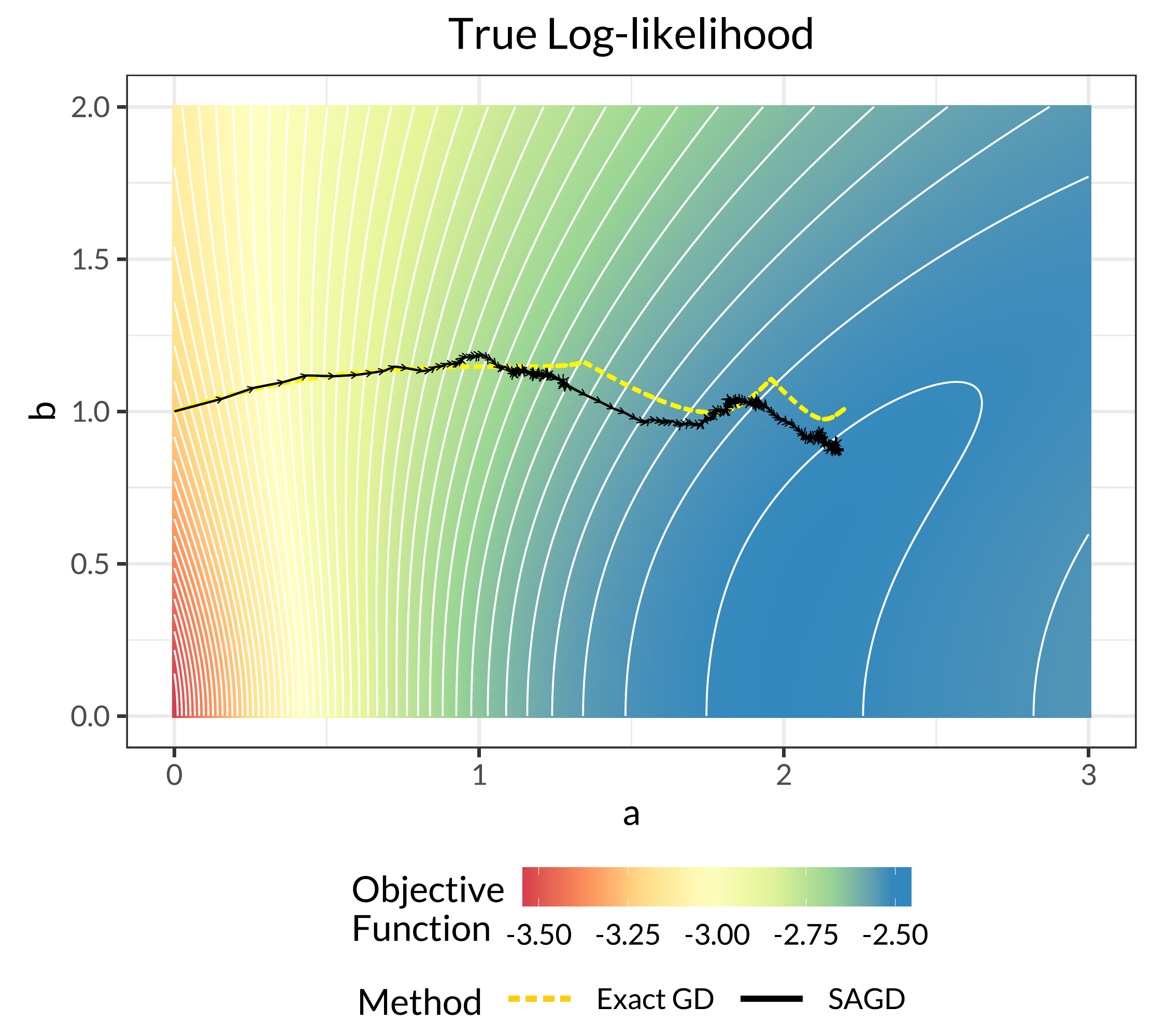}
  \caption{}
  \end{subfigure}
  \enskip{}
  \begin{subfigure}[b]{0.36\textwidth}
  \centering\includegraphics[width=\textwidth]{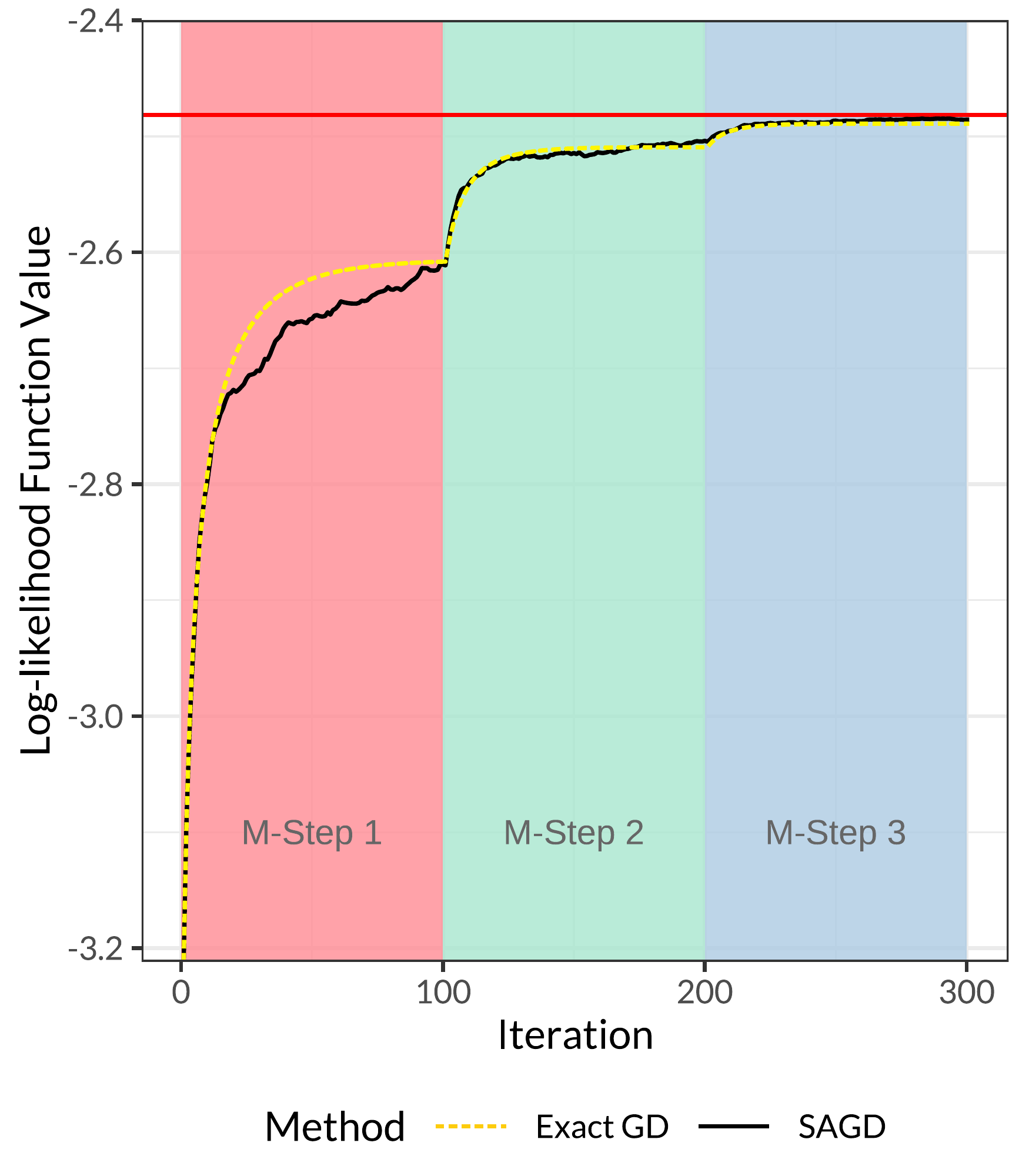}
  \caption{}
  \end{subfigure}
  \caption{\label{fig:path_loglik}(a) The paths of $(a,b)$ on the surface of
the true log-likelihood function. (b) The log-likelihood function
value versus the number of gradient updates. The horizontal line on
the top stands for the maximum log-likelihood value.}
\end{centering}
\end{figure*}

Since for this model we can evaluate the true derivatives of $Q(\theta;\theta_{k})$
using numerical integration, it is of interest to compare SAGD with
the exact gradient descent (GD) method. We set the initial value to
be $\theta_{0}=(0,1)$, and run both SAGD and exact GD for $T=100$
iterations in each M-step, with a constant step size $\alpha_{t}=0.2$.
For SAGD, Langevin parameters are specified as $\delta_{t}=0.1/\sqrt{t}$
and $K_{t}=t+20$, with the first 100 Langevin iterations discarded
as burn-in, similar to that in MCMC. Figure \ref{fig:path_loglik}(a)
demonstrates the path of $(a,b)$ values on the surface of the true
log-likelihood function after three M-steps, and Figure \ref{fig:path_loglik}(b)
gives the log-likelihood values at each gradient update. Clearly,
Figure \ref{fig:path_loglik} shows that the path of SAGD nicely approximates
that of exact GD, which further verifies the validity of the SAGD
algorithm.

\subsection{Debiased VAE}

In the second experiment, we use synthetic data to show that even
in the simplest setting, VAE can lead to biased distribution estimation,
but its bias can be effectively corrected by SAGD. The observed data
are generated as follows: given independent latent variables $Z_{i}\sim\pi(z)$,
we set $X_{i}=Z_{i}+e_{i}$, where $e_{i}\sim N(0,1)$ is independent
of $Z_{i}$, $i=1,2,\ldots,n$. The target is to recover the unknown
latent distribution $\pi(z)$ from $X_{1},\ldots,X_{n}$. In the VAE
framework, we first represent $Z$ by a deep neural network transformation
$Z=h_{\theta}(U)$, where $U\sim N(0,1)$, and then we have $p(x|u;\theta)\equiv N(h_{\theta}(u),1)$.
Once the neural network function $h_{\theta}$ has been learned, we
can simulate random variates of $Z=h_{\theta}(U)$ by generating random
$U\sim N(0,1)$, and $\pi(z)$ is approximated by the empirical distribution
of a large sample of $Z$. Therefore, by evaluating the quality of
the $Z$ sample, we can study the accuracy of the learned $h_{\theta}$
function.

\begin{figure*}[t]
\begin{centering}
\includegraphics[width=0.8\textwidth]{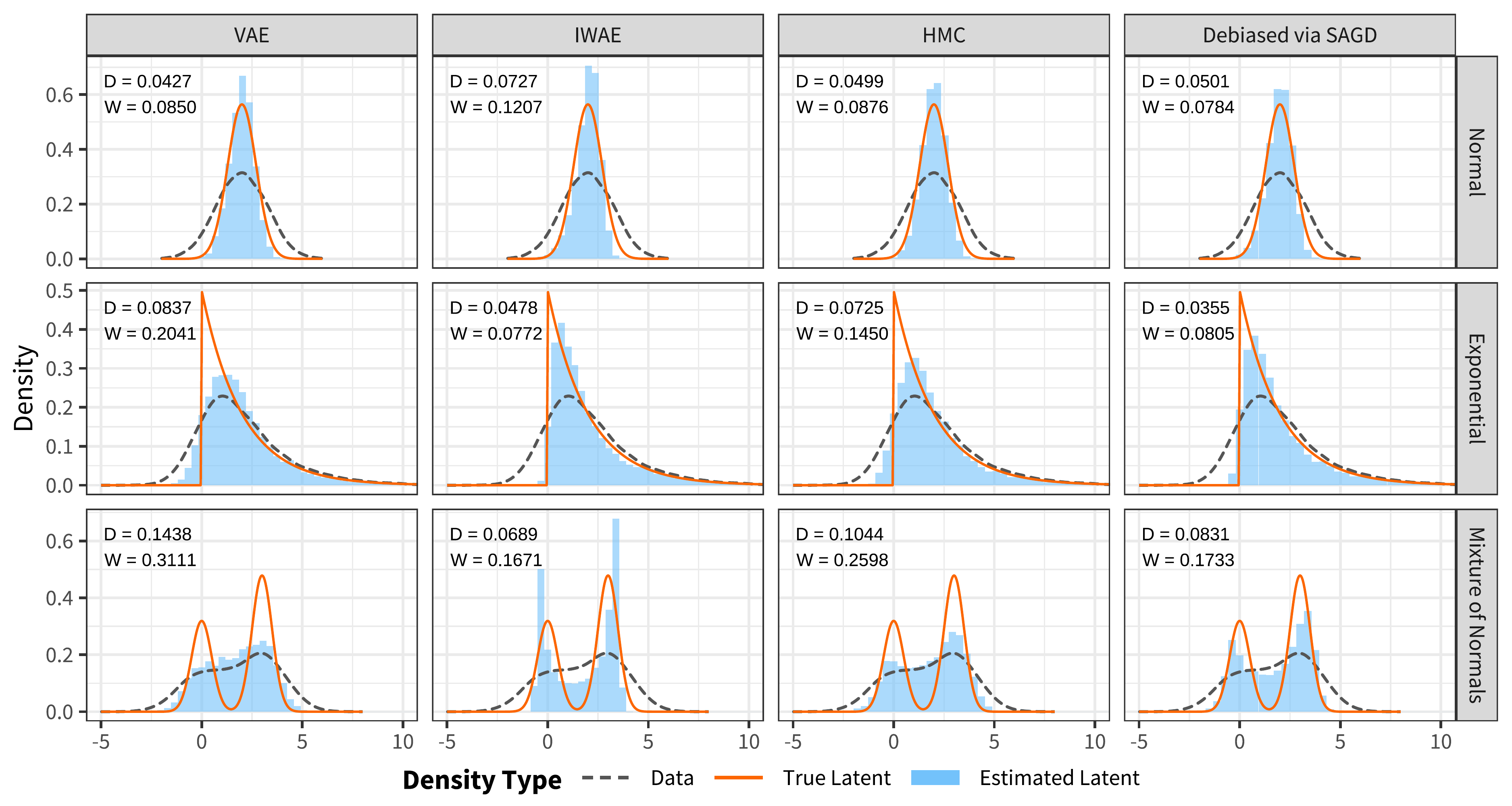}
\par\end{centering}
\caption{\label{fig:latent_estimation}A demonstration of the data distribution
(black dashed curves), true latent distribution ($\pi(z)$, red solid
curves), and estimated latent distributions ($\hat{\pi}(z)$, blue
histograms). Each row in the plot matrix corresponds to a true latent
density setting. The text on the top-left corner of each plot gives
the Kolmogorov--Smirnov distance (D) and 1-Wasserstein distance (W)
between $\pi$ and $\hat{\pi}$.}
\end{figure*}

In our experiment, we consider three true latent distributions and
generate the corresponding data sets: (a) $\pi=N(1,0.5^{2})$; (b)
an exponential distribution of mean 2; (c) a mixture of normal distributions,
$\pi=0.4\cdot N(0,0.5^{2})+0.6\cdot N(3,0.5^{2})$. For each case,
we first train a VAE model with 5000 iterations, and then fine-tune
the neural network parameter $\theta$ by running the following four
training algorithms for additional 1000 iterations: (a) VAE; (b) the
importance weighted autoencoders (IWAE, \citealp{burda2015importance})
with $k=50$ importance samples; (c) Hamiltonian Monte Carlo (HMC)
to approximate the true gradient; (d) SAGD. In HMC we use the same
step size and chain length as SAGD, and run $L=5$ leapfrog steps
to get each proposal. After training is finished, we simulate random
variates of $Z$, and compare its empirical distribution $\hat{\pi}$
with the true latent distribution $\pi$. The Kolmogorov--Smirnov
distance and 1-Wasserstein distance between $\hat{\pi}$ and $\pi$
are computed. Figure \ref{fig:latent_estimation} shows the data distribution,
true latent distribution $\pi$, and the estimated $\hat{\pi}$ in
each setting, based on one simulated data set of sample size $n=1000$. 

Figure \ref{fig:latent_estimation} reflects the following remarkable
results. For the normal case, VAE has little bias, since the true
conditional distribution $p_{U|X}(u|x)$ is indeed normal, which is
well characterized by the encoder. However, in other two cases, neither
the latent distribution $\pi(z)$ nor $p_{U|X}(u|x)$ is normal, and
hence VAE gives highly biased estimates for $\pi$. For the three
debiasing methods, the refining steps indeed reduce the bias of VAE.
However, the debiasing effect of HMC is smaller than that of SAGD,
even though theoretically they are similar. HMC also takes more computing
time due to the leapfrog steps and the calculation of acceptance probability.
For IWAE, it tends to overly truncate the support of the distribution
and exaggerate the density of modes. Overall, SAGD provides the most
favorable bias reduction results.

\begin{table}[h]
\caption{\label{tab:latent_estimation_rep}Mean and standard errors (in parentheses)
of Kolmogorov--Smirnov distance (D) and 1-Wasserstein distance (W)
between $\hat{\pi}$ and $\pi$ across replications.}

\centering{}%
\begin{tabular}{>{\raggedright}m{0.11\columnwidth}>{\raggedright}m{0.01\columnwidth}>{\centering}p{0.15\columnwidth}>{\centering}p{0.15\columnwidth}>{\centering}p{0.15\columnwidth}>{\centering}p{0.15\columnwidth}}
\toprule 
 &  & VAE & IWAE & HMC & SAGD\tabularnewline
\midrule
\midrule 
\multirow{2}{0.11\columnwidth}[-1.2em]{Normal} & \multirow{1}{0.01\columnwidth}[-0.5em]{$D$} & 0.033 (0.0024) & 0.043 (0.0035) & 0.032 (0.0019) & 0.033 (0.0020)\tabularnewline
\cmidrule{2-6} \cmidrule{3-6} \cmidrule{4-6} \cmidrule{5-6} \cmidrule{6-6} 
 & \multirow{1}{0.01\columnwidth}[-0.5em]{$W$} & 0.052 (0.0040) & 0.066 (0.0047) & 0.050 (0.0032) & 0.052 (0.0035)\tabularnewline
\midrule 
\multirow{2}{0.11\columnwidth}[-1.2em]{Exp(2)} & \multirow{1}{0.01\columnwidth}[-0.5em]{$D$} & 0.095 (0.0018) & 0.059 (0.0039) & 0.085 (0.0016) & 0.064 (0.0029)\tabularnewline
\cmidrule{2-6} \cmidrule{3-6} \cmidrule{4-6} \cmidrule{5-6} \cmidrule{6-6} 
 & \multirow{1}{0.01\columnwidth}[-0.5em]{$W$} & 0.226 (0.0079) & 0.125 (0.0091) & 0.161 (0.0051) & 0.115 (0.0082)\tabularnewline
\midrule 
\multirow{2}{0.11\columnwidth}[-1.2em]{Mixture} & \multirow{1}{0.01\columnwidth}[-0.5em]{$D$} & 0.127 (0.0027) & 0.098 (0.0061) & 0.104 (0.0015) & 0.085 (0.0019)\tabularnewline
\cmidrule{2-6} \cmidrule{3-6} \cmidrule{4-6} \cmidrule{5-6} \cmidrule{6-6} 
 & \multirow{1}{0.01\columnwidth}[-0.5em]{$W$} & 0.320 (0.0025) & 0.165 (0.0059) & 0.276 (0.0021) & 0.197 (0.0031)\tabularnewline
\midrule 
\multirow{2}{0.11\columnwidth}[-1.2em]{High-Dim.} & \multirow{1}{0.01\columnwidth}[-0.5em]{$D$} & 0.093 (0.0010) & 0.080 (0.0012) & 0.092 (0.0008) & 0.065 (0.0005)\tabularnewline
\cmidrule{2-6} \cmidrule{3-6} \cmidrule{4-6} \cmidrule{5-6} \cmidrule{6-6} 
 & \multirow{1}{0.01\columnwidth}[-0.5em]{$W$} & 0.222 (0.0027) & 0.158 (0.0029) & 0.212 (0.0030) & 0.093 (0.0015)\tabularnewline
\bottomrule
\end{tabular}
\end{table}

To further take into account the randomness in data generation, we
simulate 30 replications of the data set in each setting, and compute
the mean and standard errors of the distance metrics, shown in Table
\ref{tab:latent_estimation_rep}. Also included in this table is a
high-dimensional data set with sample size $n=10000$ and dimension
$p=100$: each dimension independently follows an exponential distribution
with mean 2. We pre-train this data set using VAE for 10000 epochs,
and then fit each method with 1000 more epochs. The mean and standard
errors are computed over all dimensions. Both Figure \ref{fig:latent_estimation}
and Table \ref{tab:latent_estimation_rep} indicate that SAGD provides
good bias reduction results for VAE in both simple and high-dimensional
settings, which highlights the importance of optimizing the correct
objective function in model fitting.

\subsection{Generative Model for MNIST Data}

\begin{figure}[h]
\begin{centering}
\includegraphics[width=0.95\columnwidth]{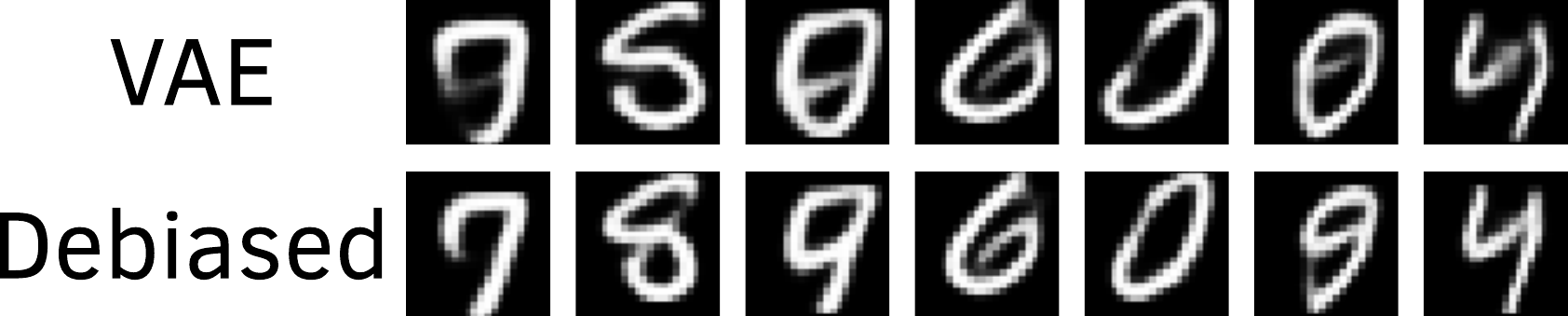}
\par\end{centering}
\caption{\label{fig:mnist}Representative examples from randomly generated
digits that show significant improvement after the refining step using
SAGD.}
\end{figure}

In the last experiment, we consider the MNIST handwritten digits data
set, and fit generative models on it. The dimension of the latent
space is set to 20, and the generative network is a combination of
convolutional filters and fully-connected layers. We first train a
VAE model for 500 epochs with a batch size of 200, and then run SAGD
for 100 epochs for fine-tuning. In SAGD, twenty independent chains
are used to compute the approximate gradient, each with five burn-in's.

Since SAGD basically refines the generative network of VAE, we can
directly compare their output images. We randomly generate 100 digits
from the trained VAE model and the debiased model, respectively, and
in Figure \ref{fig:mnist} we show some representative pairs of generated
digits, with VAE-trained ones on the top, and SAGD-refined ones on
the bottom. It is clear that the SAGD refining step improves the quality
of the generated images. For example, in the first column of Figure
\ref{fig:mnist}, VAE shows an ambiguous digit between ``9'' and
``7'', but the refined one is a definite ``7''.

\section{Conclusion}

In this article we have developed the SAGD framework for optimizing
objective functions that can be expressed as a mathematical expectation
with intractable gradients. SAGD uses the Langevin algorithm to construct
an approximate gradient in each iteration, whose accuracy is carefully
controlled. Theoretical analysis shows that SAGD has the same convergence
property as SGD, and more importantly, all the hyperparameters of
SAGD are transparent so that the algorithm can be practically implemented.
We have successfully applied SAGD to both the automated EM algorithm
and the debiased VAE. To summarize, SAGD is an alternative to the
ordinary SGD in a broader realm, and it is hoped that SAGD can be
used to solve more statistical and machine learning problems both
efficiently and reliably.

We mention two future directions for the research on SAGD. First,
one might be interested in improving the Langevin algorithm, as the
assumptions we have made are mild yet not the weakest. A second direction
is to study the convergence of SAGD combined with various acceleration
techniques, such as the momentum methods.

\appendix

\section{Appendix}

\subsection{Techincal Lemmas}

In this section we provide a number of inequalities and lemmas that
are useful for other theorems. First define the function
\[
\Gamma_{c}(x,y)=\frac{\gamma^{2}}{4}\Vert x\Vert^{2}+V(x)+\frac{\gamma}{2}\langle x,y\rangle+\frac{c}{2}\Vert y\Vert^{2}+1,
\]
where $x,y\in\mathbb{R}^{r}$ and $c>1$. Under Assumption \ref{assu:v_boundedness}(a), we
immediately obtain
\begin{align}
\Gamma_{c}(x,y) & \ge\frac{1}{6}\left\Vert \gamma x+\frac{3}{2}y\right\Vert ^{2}+\frac{\gamma^{2}}{12}\Vert x\Vert^{2}+\frac{1}{8}\Vert y\Vert^{2}+1\nonumber \\
                & \ge\frac{\gamma^{2}}{12}\Vert x\Vert^{2}+\frac{1}{8}\Vert y\Vert^{2}+1,\label{eq:lemma_gamma_c_lower_bound}
\end{align}
and
\begin{equation}
\Gamma_{c}(x,y)\le\frac{3\gamma^{2}}{8}\Vert x\Vert^{2}+V(x)+\frac{c+1}{2}\Vert y\Vert^{2}+1.\label{eq:lemma_gamma_c_upper_bound}
\end{equation}
Next, note that $V(x_{1})-V(x_{2})=\langle\nabla V(x_{2}),x_{1}-x_{2}\rangle+\int_{0}^{1}(1-s)(x_{1}-x_{2})^{\mathrm{T}}\nabla^{2}(x_{2}+s(x_{1}-x_{2}))(x_{1}-x_{2})\mathrm{d}s$.
By Assumption \ref{assu:v_boundedness}(b), we get
\begin{align}
    & \langle\nabla V(x_{2}),x_{1}-x_{2}\rangle-\frac{\nu}{2}\Vert x_{1}-x_{2}\Vert^{2}\le V(x_{1})-V(x_{2})\nonumber \\
\le & \langle\nabla V(x_{2}),x_{1}-x_{2}\rangle+\frac{\nu}{2}\Vert x_{1}-x_{2}\Vert^{2}.\label{eq:lemma_v_diff}
\end{align}
Also due to $\Vert\nabla^{2}V(x)\Vert\le\nu$, we have
\begin{equation}
\Vert\nabla V(x)\Vert\le\Vert\nabla V(0)\Vert+\Vert\nabla V(x)-\nabla V(0)\Vert\le\Vert\nabla V(0)\Vert+\nu\Vert x\Vert.\label{eq:lemma_gradv_bound}
\end{equation}

Let $\tau_{k}=(1-\gamma\delta)\rho_{k}-\delta\cdot\nabla V(\xi_{k})$,
and for $t\in[k\delta,(k+1)\delta]$ define $R_{k}(t)=\tau_{k}+\sqrt{2\gamma}(B(t)-B(k\delta))$.
Clearly $\rho_{k+1}=\tau_{k}+\sqrt{2\gamma\delta}\eta_{k}=R_{k}((k+1)\delta)$.
We then present the following two lemmas.
\begin{lem}
\label{lem:gamma_c_contraction}Let $D=\gamma^{4}C_{\beta}/\nu^{2}$, $\delta\le\min\{1/\gamma,\gamma/(2\nu),(D+1-\sqrt{D^{2}+1})/\gamma\}$,
and $c=(1-\gamma\delta/2)/(1-\gamma\delta)$.
Then there exist constants $\varepsilon=\varepsilon(\delta)>0$ and
$C_{1}=C_{1}(\delta)>0$ such that for all $k\ge0$,
\[
\Gamma_{c}(\xi_{k+1},\tau_{k})\le(1-\varepsilon\gamma\delta)\Gamma_{c}(\xi_{k},\rho_{k})+C_{1}.
\]
\end{lem}

\begin{lem}
\label{lem:gamma_c_l}Fix any integer $l>0$. Then there exists a
constant $C_{2}=C_{2}(l,\delta)>0$ such that for all $k\ge0$ and
$t\in[k\delta,(k+1)\delta]$,
\begin{align*}
& \mathbb{E}\left[\left\{ \Gamma_{c}(\xi_{k+1},R_{k}(t))\right\} ^{l}\right] \le \\
& \mathbb{E}\left[\left\{ \Gamma_{c}(\xi_{k+1},\tau_{k})\right\} ^{l}\right]+C_{2}\sum_{i=0}^{l-1}(t-k\delta)^{l-i}\mathbb{E}\left[\left\{ \Gamma_{c}(\xi_{k+1},\tau_{k})\right\} ^{i}\right].
\end{align*}
\end{lem}

For the Langevin diffusion process $W(t)$, let $\mathcal{A}$ be
its generator, defined by
\begin{align*}
(\mathcal{A}\phi)(w)= & \ \langle y,\nabla_{x}\phi(w)\rangle-\langle\nabla V(x)+\gamma y,\nabla_{y}\phi(w)\rangle \\
                      & +\gamma\mathrm{tr}(\nabla_{y}^{2}\phi(w)),
\end{align*}
where $\phi:\mathbb{R}^{2r}\to\mathbb{R}$ is any twice differentiable
function, $w=(x^{\mathrm{T}},y^{\mathrm{T}})^{\mathrm{T}}$, $\nabla_{x}\phi=\partial\phi(w)/\partial x$,
$\nabla_{y}\phi=\partial\phi(w)/\partial y$, and $\nabla_{y}^{2}\phi=\partial^{2}\phi(w)/\partial y\partial y^{\mathrm{T}}$.
Consider the functional equation $\mathcal{A}\psi=\varphi-\bar{\varphi}$,
which is called the Poisson equation, and we have the following lemma
for the solution $\psi$.
\begin{lem}
\label{lem:existence_poisson}Assume that $\varphi\in\mathscr{C}_{poly}^{r+5}$,
and the conditions in Theorem \ref{thm:finite_moments} hold. Then $\nabla^{i}\psi$, $i=0,1,2,3,4$
exist, and $\nabla^{i}\psi$ have polynomial growth.
\end{lem}

\subsection{Proof of Lemma \ref{lem:gamma_c_contraction}}

Without loss of generality consider $k=0$, and then we expand $\Gamma_{c}(\xi_{1},\tau_{0})$
using $\xi_{1}=\xi_{0}+\delta\rho_{0}$ and $\tau_{0}=(1-\gamma\delta)\rho_{0}-\delta\nabla V(\xi_{0})$.
From \eqref{eq:lemma_v_diff} and \eqref{eq:lemma_gradv_bound}, we have
\begin{align*}
& V(\xi_{1})=V(\xi_{0}+\delta\rho_{0})\le V(\xi_{0})+\delta\langle\nabla V(\xi_{0}),\rho_{0}\rangle+\frac{\nu\delta^{2}}{2}\Vert\rho_{0}\Vert^{2},\\
& \Vert\nabla V(\xi_{0})\Vert^{2}\le2\Vert\nabla V(0)\Vert^{2}+2\nu^{2}\Vert\xi_{0}\Vert^{2}.
\end{align*}
Then with some tedious calculations, it can be shown that 
\begin{align*}
      & \Gamma_{c}(\xi_{1},\tau_{0}) \\
\le\  & \Gamma_{c}(\xi_{0},\rho_{0})+\left(\frac{(2c-1)\gamma\delta(\gamma\delta-2)}{4}+\frac{\nu\delta^{2}}{2}\right)\Vert\rho_{0}\Vert^{2} \\
      & +\left(\delta-\frac{\gamma\delta^{2}}{2}-c(1-\gamma\delta)\delta\right)\langle\rho_{0},\nabla V(\xi_{0})\rangle \\
      & -\frac{\gamma\delta}{2}\langle\xi_{0},\nabla V(\xi_{0})\rangle+c\delta^{2}\left(\Vert\nabla V(0)\Vert^{2}+\nu^{2}\Vert\xi_{0}\Vert^{2}\right)+1.
\end{align*}
Setting $\delta-\gamma\delta^{2}/2-c(1-\gamma\delta)\delta=0$ yields
$c=(1-\gamma\delta/2)/(1-\gamma\delta)$. Let $\delta$ be sufficiently
small such that $\gamma\delta<1$ and $\nu\delta^{2}<\gamma\delta/2$,
\emph{i.e.}, $\delta<\min\{1/\gamma,\gamma/(2\nu)\}$. Then $c>1$
and
\begin{align*}
      & \frac{(2c-1)\gamma\delta(\gamma\delta-2)}{4}+\frac{\nu\delta^{2}}{2} \\
  =\  &-\frac{(2c-1)\gamma\delta}{2}+\frac{(2c-1)\gamma^{2}\delta^{2}}{4}+\frac{\nu\delta^{2}}{2} \\
\le\  & -\frac{(c-1)\gamma\delta}{2}.
\end{align*}
Therefore, $\Gamma_{c}(\xi_{1},\tau_{0})\le\Gamma_{c}(\xi_{0},\rho_{0})-T+C$,
where
\begin{align*}
T & =\frac{\gamma\delta}{2}\langle\xi_{0},\nabla V(\xi_{0})\rangle+\frac{(c-1)\gamma\delta}{2}\Vert\rho_{0}\Vert^{2}-c\delta^{2}\nu^{2}\Vert\xi_{0}\Vert^{2}, \\
C & =c\delta^{2}\Vert\nabla V(0)\Vert^{2}+1.
\end{align*}
Moreover, Assumption \ref{assu:v_dissipative} indicates that $\langle\nabla V(\xi_{0}),\xi_{0}\rangle/2\ge\beta V(\xi_{0})+\gamma^{2}C_{\beta}\Vert\xi_{0}\Vert^{2}-\alpha,$
so
\begin{align*}
      & \frac{\gamma\delta}{2}\langle\xi_{0},\nabla V(\xi_{0})\rangle-c\delta^{2}\nu^{2}\Vert\xi_{0}\Vert^{2} \\
\ge\  & \gamma\delta\beta V(\xi_{0})+(\gamma^{3}\delta C_{\beta}-c\delta^{2}\nu^{2})\Vert\xi_{0}\Vert^{2}-\gamma\delta\alpha.
\end{align*}
Since $c$ is decreasing in $\delta$, we further decrease $\delta$
if necessary to guarantee that $\gamma^{3}\delta C_{\beta}-c\delta^{2}\nu^{2}>\gamma^{3}\delta C_{\beta}/2$,
$i.e.$, $\delta<(D+1-\sqrt{D^{2}+1})/\gamma$, $D=\gamma^{4}C_{\beta}/\nu^{2}$.
Then
\begin{align*}
T & \ge\gamma\delta\beta V(\xi_{0})+\frac{\gamma^{3}\delta C_{\beta}}{2}\Vert\xi_{0}\Vert^{2}+\frac{(c-1)\gamma\delta}{2}\Vert\rho_{0}\Vert^{2}-\gamma\delta\alpha\\
  & =\gamma\delta\left\{ C_{\beta}\gamma^{2}\Vert\xi_{0}\Vert^{2}/2+\beta V(\xi_{0})+(c-1)\Vert\rho_{0}\Vert^{2}/2\right\} -\gamma\delta\alpha.
\end{align*}
Let $\varepsilon=\min\{4C_{\beta}/3,\beta,(c-1)/(c+1)\}$, and then
by inequality \eqref{eq:lemma_gamma_c_upper_bound}, we have $T\ge\varepsilon\gamma\delta\Gamma_{c}(\xi_{0},\rho_{0})-(\alpha+\varepsilon)\gamma\delta.$
Consequently, let $C_{1}=C+(\alpha+\varepsilon)\gamma\delta$, and
then the claimed result holds.

\subsection{Proof of Lemma \ref{lem:gamma_c_l}}

Due to the Markov property, we only need to show that for $t\in[0,\delta]$,
\begin{align}
    & \mathbb{E}\left[\left\{ \Gamma_{c}(\xi_{1},R_{0}(t))\right\} ^{l}\right]\nonumber \\
\le & \left\{ \Gamma_{c}(\xi_{1},\tau_{0})\right\} ^{l}+C_{2}\sum_{i=0}^{l-1}t^{l-i}\left\{ \Gamma_{c}(\xi_{1},\tau_{0})\right\} ^{i}.\label{eq:gamma_c_l_k0}
\end{align}

First, for $l=1$, we have $B(t)-B(0)\overset{d}{=}\sqrt{t}\zeta$,
where $\zeta\sim N(0,I_{r})$. Then
\begin{align*}
    & \Gamma_{c}(\xi_{1},R_{0}(t))-\Gamma_{c}(\xi_{1},\tau_{0}) \\
=\  & \frac{\gamma}{2}\langle\xi_{1},\sqrt{2\gamma t}\zeta\rangle+\frac{c}{2}\Vert\tau_{0}+\sqrt{2\gamma t}\zeta\Vert^{2}-\frac{c}{2}\Vert\tau_{0}\Vert^{2}\\
=\  & \frac{\gamma}{2}\langle\xi_{1},\sqrt{2\gamma t}\zeta\rangle+c\sqrt{2\gamma t}\langle\tau_{0},\zeta\rangle+c\gamma t\Vert\zeta\Vert^{2}.
\end{align*}
Using the fact that $E(\zeta)=0$ and $E(\Vert\zeta\Vert^{2})=r$,
we prove the case of $l=1$.

Then we prove the case of $l>1$ by induction. Assume that \eqref{eq:gamma_c_l_k0}
holds for all $j\le l-1$. Since $R_{0}(t)$ admits the SDE $\mathrm{d}R_{0}(t)=\sqrt{2\gamma}\mathrm{d}B(t)$,
$R_{0}(0)=\tau_{0}$, applying It\^o's formula yields 
\begin{align*}
\left\{ \Gamma_{c}(\xi_{1},R_{0}(t))\right\} ^{l}= & \left\{ \Gamma_{c}(\xi_{1},\tau_{0})\right\} ^{l}+\int_{0}^{t}\mathscr{L}G(s,R_{0}(s))\mathrm{d}s \\
& +\int_{0}^{t}\langle\nabla_{R_{0}(s)}G(s,R_{0}(s)),\sqrt{2\gamma}\mathrm{d}B(s)\rangle,
\end{align*}
where $G(\cdot,y)=\{\Gamma_{c}(\xi_{1},y)\}^{l}$, $\mathscr{L}G(\cdot,y)=\gamma\mathrm{tr}(\nabla_{y}^{2}G(\cdot,y))$,
and $\nabla_{y}G(\cdot,y)=l\{\Gamma_{c}(\xi_{1},y)\}^{l-1}(\gamma\xi_{1}/2+cy)$.
Let $v=\gamma\xi_{1}/2+cy$, $u=\Gamma_{c}(\xi_{1},y)$, and then it can be verified that
\begin{align*}
\nabla_{y}^{2}G(\cdot,y) & =clu^{l-1}I_{r}+l(l-1)u^{l-2}vv^{\mathrm{T}}, \\
\mathscr{L}G(\cdot,y) & =\gamma clru^{l-1}+\gamma l(l-1)u^{l-2}\Vert v \Vert^{2}.
\end{align*}
We also have $\Vert v \Vert^{2}\le 2c^2u$, because
\begin{align*}
 & \left\Vert \frac{\gamma}{2}x+cy\right\Vert ^{2}=2c^{2}\left(\frac{\gamma^{2}}{8c^{2}}\Vert x\Vert^{2}+\frac{\gamma}{2c}\langle x,y\rangle+\frac{1}{2}\Vert y\Vert^{2}\right)\\
\le\  & 2c^{2}\left(\frac{\gamma^{2}}{4}\Vert x\Vert^{2}+\frac{\gamma}{2}\langle x,y\rangle+\frac{c}{2}\Vert y\Vert^{2}\right)\le2c^{2}\Gamma_{c}(x,y),
\end{align*}
Finally, $\mathbb{E}\{\int_{0}^{t}\langle\nabla_{R_{0}(s)}G(s,R_{0}(s)),\sqrt{2\gamma}\mathrm{d}B(s)\rangle\}=0$,
so
\begin{align*}
& \mathbb{E}\left\{ \Gamma_{c}(\xi_{1},R_{0}(t))\right\} ^{l} \\
\le\  & \left\{ \Gamma_{c}(\xi_{1},\tau_{0})\right\} ^{l}+C\int_{0}^{t}\{\Gamma_{c}(\xi_{1},R_{0}(s))\}^{l-1}\mathrm{d}s,
\end{align*}
where $C=\gamma\left\{ clr+2l(l-1)c^2\right\}$. Using the induction hypothesis, we prove that \eqref{eq:gamma_c_l_k0}
also holds for $l$.

\subsection{Proof of Lemma \ref{lem:existence_poisson}}

Define $u(t,x,y)=\mathbb{E}(\varphi(\xi(t),\rho(t))|\xi(0)=x,\rho(0)=y)$,
and let the notation $\nabla^{i}u(t,x,y)$ denote the derivative of
$u$ with respect to $w=(x^{\mathrm{T}},y^{\mathrm{T}})^{\mathrm{T}}$.
Proposition 6.1 of \citet{kopec2015weak} shows that there exist constants
$C>0$ and $\lambda>0$ and an integer $s>0$ such that
\[
\Vert\nabla^{i}u(t,x,y)\Vert\le C(1+\Vert x\Vert^{s}+\Vert y\Vert^{s})e^{-\lambda t},\quad i=0,1,2,3,4,
\]
for all $t>0$. Moreover, it is known that $\psi$ has the representation
$\psi(w)=\int_{0}^{+\infty}u(t,x,y)\mathrm{d}t$, provided that the
integral exists. Indeed, since
\begin{align*}
& \int_{0}^{+\infty}\Vert\nabla^{i}u(t,x,y)\Vert\mathrm{d}t \\
\le  & \int_{0}^{+\infty}C(1+\Vert x\Vert^{s}+\Vert y\Vert^{s})e^{-\lambda t}\mathrm{d}t=\frac{C}{\lambda}(1+\Vert x\Vert^{s}+\Vert y\Vert^{s}),
\end{align*}
we obtain the existence of $\psi$ for $i=0$, and by the dominated
convergence theorem, we can interchange the integral and differential
operators, which shows that $\nabla^{i}\psi$, $i=1,2,3,4$ exist
and have polynomial growth.

Note that \citet{kopec2015weak} is based on a set of assumptions
\textbf{\textit{B}}-1 to \textbf{\textit{B}}-4, but only \textbf{\textit{B}}-1,
\textbf{\textit{B}}-2, and the condition $V(x)\ge0$ are used to prove
Proposition 6.1. Those three conditions are implied by Assumptions
\ref{assu:v_boundedness} and \ref{assu:v_dissipative} in this article.

\subsection{Proof of Theorem \ref{thm:finite_moments}}

In Lemma \ref{lem:gamma_c_l}, take $t=(k+1)\delta$, and then
\begin{align*}
& \mathbb{E}\left[\left\{ \Gamma_{c}(\xi_{k+1},\rho_{k+1}\right\} ^{l}\right] \\
\le\  & \mathbb{E}\left[\left\{ \Gamma_{c}(\xi_{k+1},\tau_{k})\right\} ^{l}\right]+C_{2}\sum_{i=0}^{l-1}\delta^{l-i}\mathbb{E}\left[\left\{ \Gamma_{c}(\xi_{k+1},\tau_{k})\right\} ^{i}\right].
\end{align*}
For $i=1,\ldots,l-2$, let $q_{i}=(l-1)/i>1$ and $p_{i}=q_{i}/(q_{i}-1)>1$,
and then using Young's inequality we have
\[
\delta^{l-i}\left\{ \Gamma_{c}(\xi_{k+1},\tau_{k})\right\} ^{i}\le\frac{1}{p_i}\delta^{p_{i}(l-i)}+\frac{1}{q_i}\left\{ \Gamma_{c}(\xi_{k+1},\tau_{k})\right\} ^{l-1}.
\]
As a result, there exist constants $\varepsilon_{3}=\varepsilon(l,\delta)>0$
and $C_{3}=C_{3}(l,\delta)>0$ such that
\begin{align*}
& \mathbb{E}\left[\left\{ \Gamma_{c}(\xi_{k+1},\rho_{k+1})\right\} ^{l}\right] \\
\le\  & \mathbb{E}\left[\left\{ \Gamma_{c}(\xi_{k+1},\tau_{k})\right\} ^{l}\right]+\varepsilon_{3}\mathbb{E}\left[\left\{ \Gamma_{c}(\xi_{k+1},\tau_{k})\right\} ^{l-1}\right]+C_{3}.
\end{align*}
Moreover, for any $\varepsilon_{4}>0$, we can pick a constant $C_{4}=C_{4}(\varepsilon_{3},\varepsilon_{4},C_{3})>0$
to guarantee $\varepsilon_{3}x^{l-1}+C_{3}<\varepsilon_{4}x^{l}+C_{4}$
for all $x>0$. Using these constants, we have
\begin{equation}
\mathbb{E}\left[\left\{ \Gamma_{c}(\xi_{k+1},\rho_{k+1})\right\} ^{l}\right]\le(1+\varepsilon_{4})\mathbb{E}\left[\left\{ \Gamma_{c}(\xi_{k+1},\tau_{k})\right\} ^{l}\right]+C_{4}.\label{eq:gamma_c_l_result1}
\end{equation}
With a similar argument, Lemma \ref{lem:gamma_c_contraction} indicates
that for any $\varepsilon_{5}>0$, there is a constant $C_{5}=C_{5}(\varepsilon_{5},l,\delta)>0$
such that
\begin{equation}
\left\{ \Gamma_{c}(\xi_{k+1},\tau_{k})\right\} ^{l}\le(1+\varepsilon_{5})(1-\varepsilon\gamma\delta)^{l}\left\{ \Gamma_{c}(\xi_{k},\rho_{k})\right\} ^{l}+C_{5}.\label{eq:gamma_c_l_result2}
\end{equation}
Putting \eqref{eq:gamma_c_l_result1} and \eqref{eq:gamma_c_l_result2}
together, we have
\begin{align*}
& \mathbb{E}\left[\left\{ \Gamma_{c}(\xi_{k+1},\rho_{k+1})\right\} ^{l}\right] \\
\le\  & (1+\varepsilon_{4})(1+\varepsilon_{5})(1-\varepsilon\gamma\delta)^{l}\mathbb{E}\left[\left\{ \Gamma_{c}(\xi_{k},\rho_{k})\right\} ^{l}\right]+(1+\varepsilon_{4})C_{5}+C_{4}.
\end{align*}
Clearly, by choosing $\varepsilon_{4}$ and $\varepsilon_{5}$ such
that $r_{l}=(1+\varepsilon_{4})(1+\varepsilon_{5})(1-\varepsilon\gamma\delta)^{l}<1$,
we get 
\[
\mathbb{E}\left[\left\{ \Gamma_{c}(\xi_{k},\rho_{k})\right\} ^{l}\right]\le r_{l}^{k}\left\{ \Gamma_{c}(\xi_{0},\rho_{0})\right\} ^{l}+\{(1+\varepsilon_{4})C_{5}+C_{4}\}\sum_{i=0}^{k-1}r_{l}^{i}.
\]
Formula \eqref{eq:lemma_gamma_c_lower_bound} shows that $\left\{ \Gamma_{c}(\xi_{k},\rho_{k})\right\} ^{l}$
is lower bounded by by a polynomial of order $2l$, and Assumption
\ref{assu:v_boundedness}(c) and \eqref{eq:lemma_gamma_c_upper_bound} show that $\left\{ \Gamma_{c}(\xi_{0},\rho_{0})\right\} ^{l}$
is upper bounded by a polynomial. Then the desired result is proved.

\subsection{Proof of Theorem \ref{thm:control_bias_mse}}

The proof of this theorem is inspired by \citet{mattingly2010convergence}
and \citet{vollmer2016exploration}. First let $U(W_{k})=(\rho_{k}^{\mathrm{T}},-\gamma\rho_{k}^{\mathrm{T}}-\nabla V(\xi_{k})^{\mathrm{T}})^{\mathrm{T}}$
and $\zeta_{k}=(\mathbf{0}^{\mathrm{T}},\eta_{k}^{\mathrm{T}})^{\mathrm{T}}$,
and then we have the relation $W_{k+1}=W_{k}+\delta U(W_{k})+\sqrt{2\gamma\delta}\zeta_{k}$.
For simplicity denote $\varphi_{k}=\varphi(W_{k})$, $u_{k}=U(W_{k})$,
$\psi_{k}=\psi(W_{k})$, and $d_{k}=W_{k+1}-W_{k}$. We also use $\nabla^{i}\psi_{k}$
to denote the $i$-th derivative of $\psi$ at $W_{k}$, and $\nabla^{i}\psi(x)[d,\ldots,d]$
to denote the derivative $\nabla^{i}\psi(x)$ evaluated in the directions
$(d,\ldots,d)$. In addition, $\nabla_{y}^{i}\psi_{k}$ stands for
the partial derivative of $\psi$ at $W_{k}$ with respect to the
second component. Since $\psi$ has a fourth-order derivative, the
following Taylor expansion holds,
\begin{align}
\psi_{k+1}= & \psi_{k}+\langle d_{k},\nabla\psi_{k}\rangle+\frac{1}{2}\langle d_{k},(\nabla^{2}\psi_{k})d_{k}\rangle\nonumber \\
            & +\frac{1}{6}\nabla^{3}\psi_{k}[d_{k},d_{k},d_{k}]+R_{k+1},\label{eq:taylor_psi}
\end{align}
where
\begin{align*}
& R_{k+1} \\
=\  & \frac{1}{6}\left(\int_{0}^{1}s^{3}\nabla^{4}\psi(sW_{k}+(1-s)W_{k+1})\mathrm{d}s\right)[d_{k},d_{k},d_{k},d_{k}]
\end{align*}
is the remainder term. By expanding $d_{k}=\delta u_{k}+\sqrt{2\gamma\delta}\cdot\zeta_{k}$
and using the definition of the generator $\mathcal{A}$, we can show
that
\begin{align*}
\langle d_{k},\nabla\psi_{k}\rangle & =\delta\langle u_{k},\nabla\psi_{k}\rangle+\sqrt{2\gamma\delta}\cdot\langle\zeta_{k},\nabla\psi_{k}\rangle\\
 & =\delta\mathcal{A}\psi_{k}-\gamma\delta\mathrm{tr}(\nabla_{y}^{2}\psi_{k})+\sqrt{2\gamma\delta}\cdot\langle\zeta_{k},\nabla\psi_{k}\rangle\\
 & =\delta(\varphi_{k}-\bar{\varphi})-\gamma\delta\mathrm{tr}(\nabla_{y}^{2}\psi_{k})+\sqrt{2\gamma\delta}\cdot\langle\zeta_{k},\nabla\psi_{k}\rangle
\end{align*}
and
\begin{align*}
\frac{1}{2}\langle d_{k},(\nabla^{2}\psi_{k})d_{k}\rangle=\  & \frac{1}{2}\delta^{2}\langle u_{k},(\nabla^{2}\psi_{k})u_{k}\rangle \\
& +\sqrt{2\gamma\delta^{3}}\cdot\langle\zeta_{k},(\nabla^{2}\psi_{k})u_{k}\rangle \\
& +\gamma\delta\langle\xi_{k},(\nabla_{y}^{2}\psi_{k})\xi_{k}\rangle.
\end{align*}
Since $\xi_{k}\sim N(0,I_{r})$ and $\xi_{k}$ is independent of $W_{k}$,
we have for all $k$, $\mathbb{E}\langle\zeta_{k},\nabla\psi_{k}\rangle=0$,
$\mathbb{E}\langle\zeta_{k},(\nabla^{2}\psi_{k})u_{k}\rangle=0$,
$\mathbb{E}\langle\xi_{k},(\nabla_{y}^{2}\psi_{k})\xi_{k}\rangle=\mathrm{tr}(\nabla_{y}^{2}\psi_{k})$
etc.. Therefore, taking the expectation on both sides of \eqref{eq:taylor_psi}
cancels many terms involving $\xi_{k}$. Let the notation $X\overset{E}{=}Y$
stand for $\mathbb{E}(X)=\mathbb{E}(Y)$, and then after some simplification,
we get
\begin{align}
\psi_{k+1}\overset{E}{=}\  & \psi_{k}+\delta(\varphi_{k}-\bar{\varphi})+\frac{\delta^{2}}{2}\langle u_{k},(\nabla^{2}\psi_{k})u_{k}\rangle\nonumber \\
& +\gamma\delta^{2}\nabla^{3}\psi_{k}[u_{k},e,e]+\frac{\delta^{3}}{6}\nabla^{3}\psi_{k}[u_{k},u_{k},u_{k}]\nonumber \\
& +R_{k+1},\label{eq:psi_expectation}
\end{align}
where $e=(\mathbf{0}^{\mathrm{T}},\mathbf{1}^{\mathrm{T}})^{\mathrm{T}}$.
Summing \eqref{eq:psi_expectation} over the first $K$ terms, and
dividing both sides by $K\delta$, we obtain
\begin{align}
& \frac{1}{K\delta}(\psi_{K}-\psi_{0})\nonumber \\
\overset{E}{=}\  & \hat{\varphi}-\bar{\varphi}+\frac{\delta}{2K}\sum_{k=0}^{K-1}\langle u_{k},(\nabla^{2}\psi_{k})u_{k}\rangle+\frac{\gamma\delta}{K}\sum_{k=0}^{K-1}\nabla^{3}\psi_{k}[u_{k},e,e]\nonumber \\
 & +\frac{\delta^{2}}{6K}\sum_{k=0}^{K-1}\nabla^{3}\psi_{k}[u_{k},u_{k},u_{k}]+\frac{1}{K\delta}\sum_{k=0}^{K-1}R_{k+1}.\label{eq:psi_average}
\end{align}

Now we attempt to bound each term in \eqref{eq:psi_average}. First,
$|\psi_{K}-\psi_{0}|\le|\psi_{0}|+|\psi_{K}|$, so by Lemma \ref{lem:existence_poisson}
we know that it is bounded by a polynomial of $W_{K}$. Then Theorem
\ref{thm:finite_moments} indicates that the expectation $\mathbb{E}|\psi_{K}-\psi_{0}|$
is bounded by a constant, denoted by $A_{1}$. Next, $\vert\langle u_{k},(\nabla^{2}\psi_{k})u_{k}\rangle\vert\le\Vert\nabla^{2}\psi_{k}\Vert\cdot\Vert v_{k}\Vert^{2}$,
which can be bounded by a product of polynomials of $W_{k}$. Using
the same argument, we get $\mathbb{E}\vert\langle u_{k},(\nabla^{2}\psi_{k})u_{k}\rangle\vert\le A_{2}$
for some constant $A_{2}$. Similar analysis for higher order terms
shows $\mathbb{E}\vert\nabla^{3}\psi_{k}[u_{k},e,e]\vert\le A_{3}$
and $\mathbb{E}\vert\nabla^{3}\psi_{k}[u_{k},u_{k},u_{k}]\vert\le A_{4}$.
For the remainder term, since $d_{k}=\sqrt{\delta}(\sqrt{\delta}u_{k}+\sqrt{2\gamma}\zeta_{k})$,
we have $\mathbb{E}\vert R_{k+1}\vert\le\delta^{2}A_{5}$ for some
constant $A_{5}$. Combining all these terms together, we eventually
get the first inequality.

For the second part, let $T=K\delta$, and then from \eqref{eq:taylor_psi}
we get
\begin{align*}
 & \frac{1}{T}(\psi_{K}-\psi_{0}) \\
=\  & \hat{\varphi}-\bar{\varphi}+\frac{\gamma\delta}{T}\sum_{k=0}^{K-1}\left\{ \langle\xi_{k},(\nabla_{y}^{2}\psi_{k})\xi_{k}\rangle-\mathrm{tr}(\nabla_{y}^{2}\psi_{k})\right\} \\
 & +\frac{\sqrt{2\gamma\delta}}{T}\sum_{k=0}^{K-1}\langle\zeta_{k},\nabla\psi_{k}\rangle\\
 & +\frac{\sqrt{2\gamma\delta^{3}}}{T}\sum_{k=0}^{K-1}\langle\zeta_{k},(\nabla^{2}\psi_{k})u_{k}\rangle+\frac{\delta^{2}}{2T}\sum_{k=0}^{K-1}\langle u_{k},(\nabla^{2}\psi_{k})u_{k}\rangle\\
 & +\frac{1}{6T}\sum_{k=0}^{K-1}\nabla^{3}\psi_{k}[d_{k},d_{k},d_{k}]+\frac{1}{T}\sum_{k=0}^{K-1}R_{k+1}\\
\coloneqq\  & \hat{\varphi}-\bar{\varphi}+\frac{\gamma\delta}{T}S_{1}+\frac{\sqrt{2\gamma\delta}}{T}S_{2}+\frac{\sqrt{2\gamma\delta^{3}}}{T}S_{3}+\frac{\delta^{2}}{2T}S_{4} \\
 & +\frac{1}{6T}S_{5}+\frac{1}{T}S_{6}.
\end{align*}
Therefore,
\begin{align*}
\mathbb{E}(\hat{\varphi}-\bar{\varphi})^{2}\le\  & \frac{7}{T^{2}}\mathbb{E}\bigg\{ (\psi_{K}-\psi_{0})^{2}+\gamma^{2}\delta^{2}S_{1}^{2}+2\gamma\delta S_{2}^{2} \\
& \left . +2\gamma\delta^{3}S_{3}^{2}+\frac{\delta^{4}}{4}S_{4}^{2}+\frac{1}{36}S_{5}^{2}+S_{6}^{2}\right\} .
\end{align*}
Let $B_{0},\ldots,B_{6}$ represent some positive constants. Similar
to the first part, we first show $\mathbb{E}(\psi_{K}-\psi_{0})^{2}\le B_{0}$.
Then note that $S_{1}$ is a martingale, so
\begin{align*}
\mathbb{E}(S_{1}^{2}) & =\sum_{k=0}^{K-1}\mathbb{E}\left\{ \langle\xi_{k},(\nabla_{y}^{2}\psi_{k})\xi_{k}\rangle-\mathrm{tr}(\nabla_{y}^{2}\psi_{k})\right\} ^{2}\le B_{1}K.
\end{align*}
With analogous calculations, we can verify that $\mathbb{E}(S_{2}^{2})\le B_{2}K$,
$\mathbb{E}(S_{3}^{2})\le B_{3}K$, $\mathbb{E}(S_{4}^{2})\le B_{4}K^{2}$,
$\mathbb{E}(S_{5}^{2})\le B_{5}(\delta^{4}K^{2}+\delta^{3}K)$, and
$\mathbb{E}(S_{6}^{2})\le B_{6}\gamma^{4}K^{2}$. Then adding up the
terms gives the desired result.

\subsection{Proof of Corollary \ref{cor:nn}}

It is easy to show that $V(z)=-\log p(x,z)+C$, where $p(x,z)$ is
the joint density of $(X,Z)$ and $C$ is free of $z$. For simplicity
we let $\sigma=1$, since it only affects constant terms or scaling
factors. Let $y=Wz+b=(y_{1},\ldots,y_{m})^{\mathrm{T}}$, and then
\[
V(z)=-\log p(x,z)+C=\frac{1}{2}\sum_{i=1}^{m}\{a(y_{i})-x_{i}\}^{2}+\frac{1}{2}\Vert z\Vert^{2}+C.
\]
Since the activation function $a(x)=\log(1+e^{x})$ is smooth, Assumption
1 trivially holds.

Define $l_{1}(y)=\max(y,0)-1$ and $l_{2}(y)=\max(y,0)+1$, then clearly
$l_{1}(y)<a(y)<l_{2}(y)$ for all $y\in\mathbb{R}$. Therefore, $\{a(y_{i})-x_{i}\}^{2}$
must be bounded above by a quadratic function of $y_{i}$. As $y$
is a linear transformation of $z$, it is also true that $V(z)$ is
bounded above by a quadratic function of $z$: $V(z)\le C_{1}(1+\Vert z\Vert^{2})$
for some $C_{1}\ge0$.

On ther other hand, $\partial V/\partial y_{i}=\{a(y_{i})-x_{i}\}a'(y_{i})$,
where $a'(y)=e^{y}/(1+e^{y})$. Let $u=(\partial V/\partial y_{1},\ldots,\partial V/\partial y_{m})^{\mathrm{T}}$,
and then $\nabla V(z)=W^{\mathrm{T}}u+z$ and
\[
\langle\nabla V(z),z\rangle=u^{\mathrm{T}}Wz+z^{\mathrm{T}}z=u^{\mathrm{T}}(y-b)+\Vert z\Vert^{2}.
\]
We can show that $0<a'(y)<1$, $l_{1}(y)<a(y)a'(y)<l_{2}(y)$, $l_{1}(y)<a'(y)y<l_{2}(y)$,
and $a(y)a'(y)y>-1$. So
\begin{align*}
u_{i}(y_{i}-b_{i})=\  & a(y_{i})a'(y_{i})y_{i}-x_{i}a'(y_{i})y_{i} \\
& -b_{i}a(y_{i})a'(y_{i})-b_{i}x_{i}a'(y_{i})
\end{align*}
is bounded below by a piecewise linear function of $y$, which is
also a piecewise linear function of $z$. Consequently, $\langle\nabla V(z),z\rangle$
is bounded below by a quadratic function of $z$: $\langle\nabla V(z),z\rangle\ge C_{2}\Vert z\Vert^{2}-C_{3}$
for some $C_{2},C_{3}\ge0$. Combining with the upper bound of $V(z)$,
we show that Assumption 2 holds with a sufficiently small $\beta$
and a sufficiently large $\alpha$.

\subsection{Proof of Theorem \ref{thm:convergence_convex}}

The Lipschitz continuity of $F$ implies that $\Vert g(\theta)\Vert=\Vert\nabla F(\theta)\Vert\le L$
for all $\theta\in\Theta$. Let $F^{*}=\min_{\theta\in\Theta}\,F(\theta)$,
and $\theta^{*}\in\arg\min_{\theta\in\Theta}\,F(\theta)$. Then by
the convexity of $F$ we have $F(\theta)-F^{*}\le\langle g(\theta),\theta-\theta^{*}\rangle$
for all $\theta\in\Theta$.

In what follows $\mathbb{E}_{t}[\cdot]$ denotes the expectation with
respect to $\{\xi_{t,k}\}_{k=0}^{K_{t}-1}$, and $\mathbb{E}[\cdot]$
is the total expectation. When no confusion is caused, we write
$g\equiv g(\theta_{t})$ and $\tilde{g}_{t}\equiv\tilde{g}_{t}(\theta_{t})$ for brevity.
Suppose that $\Vert\mathbb{E}_{t}[\tilde{g}_{t}]-g\Vert\le\varepsilon_{t}$
and $\mathbb{E}_{t}\Vert\tilde{g}_{t}-g\Vert^{2}\le\omega_{t}$,
and then we have
\begin{align*}
\quad\mathbb{E}_{t}\langle\tilde{g}_{t}-g,\theta_{t}-\theta^{*}\rangle & \ge_{(i)}-\Vert\mathbb{E}_{t}[\tilde{g}_{t}]-g\Vert\cdot\Vert\theta_{t}-\theta^{*}\Vert\\
 & \ge-\varepsilon_{t}\Vert\theta_{t}-\theta^{*}\Vert\ge-\varepsilon_{t}D
\end{align*}
and
\begin{align*}
\quad\mathbb{E}_{t}\Vert\tilde{g}_{t}\Vert^{2} & =\mathbb{E}_{t}\left[\Vert\tilde{g}_{t}-g\Vert^{2}+2\langle\tilde{g}_{t}-g,g\rangle+\Vert g\Vert^{2}\right]\\
 & \le_{(ii)}\omega_{t}+2\Vert\mathbb{E}_{t}[\tilde{g}_{t}]-g\Vert\cdot\Vert g\Vert+L^{2}\\
 & \le\omega_{t}+2\varepsilon_{t}L+L^{2},
\end{align*}
where $(i)$ and $(ii)$ use the Cauchy--Schwarz inequality. Therefore,
\begin{align*}
\mathbb{E}_{t}\langle\tilde{g}_{t},\theta_{t}-\theta^{*}\rangle & =\mathbb{E}_{t}\langle\tilde{g}_{t}-g,\theta_{t}-\theta^{*}\rangle+\langle g,\theta_{t}-\theta^{*}\rangle\\
 & \ge F(\theta_t)-F^{*}-\varepsilon_{t}D,
\end{align*}
and the update formula for $\theta_{t+1}$ indicates that
\begin{align*}
 & \,\,\quad\mathbb{E}_{t}\Vert\theta_{t+1}-\theta^{*}\Vert^{2} \\
 & \le_{(iii)}\mathbb{E}_{t}\Vert\theta_{t}-\alpha_{t}\cdot\tilde{g}_{t}-\theta^{*}\Vert^{2}\\
 & \le\Vert\theta_{t}-\theta^{*}\Vert^{2}-2\alpha_{t}\mathbb{E}_{t}\langle\tilde{g}_{t},\theta_{t}-\theta^{*}\rangle+\alpha_{t}^{2}\mathbb{E}_{t}\Vert\tilde{g}_{t}\Vert^{2}\\
 & \le\Vert\theta_{t}-\theta^{*}\Vert^{2}-2\alpha_{t}(F(\theta_{t})-F^{*})+2\alpha_{t}\varepsilon_{t}D \\
 & \quad +\alpha_{t}^{2}(\omega_{t}+2\varepsilon_{t}L+L^{2}),
\end{align*}
where $(iii)$ comes from the nonexpansion property of the projection
operator. Reorganizing the inequality above yields
\begin{equation}
F(\theta_{t})-F^{*}\le\frac{\Vert\theta_{t}-\theta^{*}\Vert^{2}}{2\alpha_{t}}-\frac{\mathbb{E}_{t}\Vert\theta_{t+1}-\theta^{*}\Vert^{2}}{2\alpha_{t}}+\mu_{t},\label{eq:telescope}
\end{equation}
where $\mu_{t}=\varepsilon_{t}D+\alpha_{t}(\omega_{t}+2\varepsilon_{t}L+L^{2})/2$.
Summarizing \eqref{eq:telescope} over $t=1,2,\ldots,T$ and taking
the total expectation, we obtain
\begin{align*}
& \mathbb{E}\sum_{t=1}^{T}\{F(\theta_{t})-F^{*}\} \\
\le\  & \frac{\mathbb{E}\Vert\theta_{1}-\theta^{*}\Vert^{2}}{2\alpha_{1}}+\sum_{t=2}^{T}\left(\frac{1}{2\alpha_{t}}-\frac{1}{2\alpha_{t-1}}\right)\mathbb{E}\Vert\theta_{t}-\theta^{*}\Vert^{2}+\sum_{t=1}^{T}\mu_{t}.
\end{align*}
Take $\alpha_{t}=\alpha_{0}/\sqrt{t}$, so $(2\alpha_{t})^{-1}-(2\alpha_{t-1})^{-1}>0$,
and hence
\begin{align*}
\mathbb{E}\sum_{t=1}^{T}\{F(\theta_{t})-F^{*}\} & \le\frac{D^{2}}{2\alpha_{1}}+D^{2}\sum_{t=2}^{T}\left(\frac{1}{2\alpha_{t}}-\frac{1}{2\alpha_{t-1}}\right)+\sum_{t=1}^{T}\mu_{t}\\
& =\frac{D^{2}\sqrt{T}}{2\alpha_{0}}+\sum_{t=1}^{T}\mu_{t}.
\end{align*}
Choose $\delta_{t}=1/\sqrt{t}$ and $K_{t}=t$, and then $\varepsilon_{t}=2C_{1}/\sqrt{t}$
and $\omega_{t}=C_{2}(1/\sqrt{t}+1/t)$. Consequently, we see that
$\mu_{t}=\mathcal{O}(1/\sqrt{t})$. Since $\sum_{t=1}^{T}1/\sqrt{t}\le2\sqrt{T}$,
we conclude that $\mathbb{E}\sum_{t=1}^{T}\{F(\theta_{t})-F^{*}\}\le\mathcal{O}(\sqrt{T})$.
Finally, by the convexity of $F$ we have $T^{-1}\sum_{t=1}^{T}F(\theta_{t})\ge F(\hat{\theta})$,
and then the proof is complete.

\subsection{Proof of Theorem \ref{thm:convergence_nonconvex}}

Similar to the proof of Theorem \ref{thm:convergence_convex}, denote
$g\equiv g(\theta_{t})$ and $\tilde{g}_{t}\equiv\tilde{g}_{t}(\theta_{t})$.
Suppose that $\Vert\mathbb{E}_{t}[\tilde{g}_{t}]-g\Vert\le\varepsilon_{t}$
and $\mathbb{E}_{t}\Vert\tilde{g}_{t}-g\Vert^{2}\le\omega_{t}$,
and then we have
\begin{align*}
\mathbb{E}_{t}\langle\tilde{g}_{t},g\rangle & =\mathbb{E}_{t}\langle\tilde{g}_{t}-g,g\rangle+\Vert g\Vert^{2}\\
 & \ge-\frac{1}{2}\left\{ \Vert\mathbb{E}_{t}[\tilde{g}_{t}]-g\Vert^{2}+\Vert g\Vert^{2}\right\} +\Vert g\Vert^{2}\\
 & \ge\frac{1}{2}\Vert g\Vert^{2}-\frac{1}{2}\varepsilon_{t}^{2},\\
\mathbb{E}_{t}\Vert\tilde{g}_{t}\Vert^{2} & \le2\mathbb{E}_{t}\left[\Vert\tilde{g}_{t}-g\Vert^{2}+\Vert g\Vert^{2}\right]\le2\omega_{t}+2\Vert g\Vert^{2}.
\end{align*}
It is well known that if $g(\theta)\coloneqq\nabla F(\theta)$ is
$G$-Lipschitz continuous, then for any $\theta'$ and $\theta$,
\[
F(\theta')-F(\theta)\le\langle g(\theta),\theta'-\theta\rangle+\frac{1}{2}G\Vert\theta'-\theta\Vert^{2}.
\]
Therefore,
\begin{align}
 & \mathbb{E}_{t}[F(\theta_{t+1})]-F(\theta_{t})\nonumber \\
\le\  & \mathbb{E}_{t}\langle g,\theta_{t+1}-\theta_{t}\rangle+\frac{G}{2}\mathbb{E}_{t}\Vert\theta_{t+1}-\theta_{t}\Vert^{2}\nonumber \\
=\  & -\alpha_{t}\mathbb{E}_{t}\langle g,\tilde{g}\rangle+\frac{1}{2}\alpha_{t}^{2}G\mathbb{E}_{t}\Vert\tilde{g}_{t}\Vert^{2}\nonumber \\
\le\  & \frac{1}{2}\alpha_{t}\varepsilon_{t}^{2}-\frac{\alpha_{t}}{2}\Vert g\Vert^{2}+\alpha_{t}^{2}\omega_{t}G+\alpha_{t}^{2}G\Vert g\Vert^{2}\nonumber \\
=\  & -\frac{1}{2}\alpha_{t}(1-2\alpha_{t}G)\Vert g\Vert^{2}+\frac{1}{2}\alpha_{t}\varepsilon_{t}^{2}+\alpha_{t}^{2}\omega_{t}G.\label{eq:f_decrease}
\end{align}
Let $\beta_{t}=\alpha_{t}(1-2\alpha_{t}G)$, and then take the total
expectation on both sides of \eqref{eq:f_decrease}, yielding
\[
\mathbb{E}[F(\theta_{t+1})]-\mathbb{E}[F(\theta_{t})] \le-\frac{1}{2}\beta_{t}\mathbb{E}[\Vert g\Vert^{2}]+\frac{1}{2}\alpha_{t}\varepsilon_{t}^{2}+\alpha_{t}^{2}\omega_{t}G
\]
and
\begin{align*}
\sum_{t=1}^{T}\beta_{t}\mathbb{E}[\Vert g\Vert^{2}] & \le2\mathbb{E}[F(\theta_{1})-F(\theta_{T+1})]+\alpha_{t}\varepsilon_{t}^{2}+2\alpha_{t}^{2}\omega_{t}G\\
 & \le2\mathbb{E}[F(\theta_{1})]-2F^{*}+\alpha_{t}\varepsilon_{t}^{2}+2\alpha_{t}^{2}\omega_{t}G,
\end{align*}
where $F^{*}$ is the optimal value. The choice of $\{\alpha_{t}\}$,
$\{\varepsilon_{t}\}$, and $\{\omega_{t}\}$ in the theorem guarantees
that $\sum_{t=1}^{\infty}\beta_{t}=\infty$ and $\sum_{t=1}^{\infty}\beta_{t}\mathbb{E}[\Vert g\Vert^{2}]<\infty$,
so the conclusion holds.

\bibliographystyle{aaai}
\bibliography{ref}

\end{document}